\documentclass[pdflatex,sn-mathphys]{sn-jnl}% Math and Physical Sciences Reference Style
\jyear{2021}%

\theoremstyle{thmstyleone}%
\theoremstyle{thmstyletwo}%

\theoremstyle{thmstylethree}%

\usepackage{mathptmx} % important ijcv package to control fonts.
\usepackage{makecell}
\usepackage[final]{microtype}
\usepackage[USenglish]{babel}
\usepackage{etoolbox}
\usepackage[T1]{fontenc}
\usepackage{amsmath}
\usepackage{amssymb}
\usepackage{natbib}
\usepackage{makecell}
\usepackage{afterpage}
\usepackage{placeins}
\usepackage{pifont} % \ding symbols
\usepackage{threeparttable}
\usepackage{epigraph}
\usepackage{multirow}
\usepackage{tabularx}
\usepackage{dsfont}
\usepackage{url}
\usepackage[export]{adjustbox}
\usepackage{booktabs}

\usepackage{subcaption} % Felix: use this one
\usepackage[font=small,labelfont=bf]{caption}
\usepackage{longtable}
\usepackage{tabu}
\usepackage{supertabular}
\usepackage{enumitem}

\newcommand{\figref}[1]{Fig.~\ref{#1}}
\newcommand{\reqref}[1]{Eq.~\eqref{#1}}
\newcommand{\secref}[1]{Sec.~\ref{#1}}
\newcommand{\tableref}[1]{Table~\ref{#1}}

\usepackage{xspace}
\makeatletter
\DeclareRobustCommand\onedot{\futurelet\@let@token\@onedot}
\def\@onedot{\ifx\@let@token.\else.\null\fi\xspace}
\def\eg{\emph{e.g}\onedot} 
\def\ie{\emph{i.e}\onedot} 
 
\def\etc{\emph{etc}\onedot} 
\def\wrt{w.r.t\onedot} 
\def\etal{\emph{et al}\onedot}
\makeatother

\newcommand{\topone}[1]{\textbf{\textbf{#1}}}
\newcommand{\toptwo}[1]{#1}

\usepackage{hyperref}
\begin{document}

\title[\textsc{Sparta}: Spatially Attentive and Adversarially Robust Activation]{\textsc{Sparta}: Spatially Attentive and Adversarially Robust Activation}


\author[1]{\fnm{Qing} \sur{Guo}} 
% \email{iauthor@gmail.com}

\author[2]{\fnm{Felix} \sur{Juefei-Xu}}
% \email{iiauthor@gmail.com}

\author[3]{\fnm{Changqing} \sur{Zhou}}
% \email{iiiauthor@gmail.com}

\author[4]{\fnm{Wei} \sur{Feng}}
% \email{iiiauthor@gmail.com}

\author[3]{\fnm{Yang} \sur{Liu}}
% \email{iiiauthor@gmail.com}

\author[5]{\fnm{Song} \sur{Wang}}
% \email{iiiauthor@gmail.com}

\affil[1]{
% \orgdiv{Department}, 
\orgname{Centre for Frontier AI Research, A*STAR},
\orgaddress{
%
% \street{Street}, \city{City}, \postcode{100190}, \state{State},
%
\country{Singapore}}
}
\affil[2]{
% \orgdiv{Department}, 
\orgname{New York University},
\orgaddress{
%
% \street{Street}, \city{City}, \postcode{100190}, \state{State},
%
\country{USA}}
}
\affil[3]{
% \orgdiv{Department}, 
\orgname{Nanyang Technological University},
\orgaddress{
%
% \street{Street}, \city{City}, \postcode{100190}, \state{State},
%
\country{Singapore}}
}
\affil[4]{
% \orgdiv{Department}, 
\orgname{Tianjin University},
\orgaddress{
%
% \street{Street}, \city{City}, \postcode{100190}, \state{State},
%
\country{China}}
}
\affil[5]{
% \orgdiv{Department of Computer Science and Engineering}, 
\orgname{University of South Carolina},
\orgaddress{
%
% \street{Street}, \city{City}, \postcode{100190}, \state{State},
%
\country{USA}}
}


\abstract{
Adversarial training (AT) is one of the most effective ways for improving the robustness of deep convolution neural networks (CNNs). Just like common network training, the effectiveness of AT relies on the design of basic network components. In this paper, we conduct an in-depth study on the role of the basic ReLU activation component in AT for robust CNNs. We find that the spatially-shared and input-independent properties of ReLU activation make CNNs less robust to white-box adversarial attacks with either standard or adversarial training. To address this problem, we extend ReLU to a novel \textsc{Sparta} activation function (Spatially attentive and  Adversarially Robust Activation), which enables CNNs to achieve higher robustness, \ie, lower error rate on adversarial examples than the existing state-of-the-art (SOTA) activation functions, without hurting accuracy on clean examples. We further study the relationship between \textsc{Sparta} and the SOTA activation functions, providing more insights about the advantages of our method. With comprehensive experiments, we also find that the proposed method exhibits superior cross-CNN and cross-dataset transferability. For the former, the adversarially trained \textsc{Sparta} function for one CNN (\eg, ResNet-18) can be fixed and directly used to train another adversarially robust CNN (\eg, ResNet-34). For the latter, the \textsc{Sparta} function trained on one dataset (\eg, CIFAR-10) can be employed to train adversarially robust CNNs on another dataset (\eg, SVHN). In both cases, \textsc{Sparta} leads to CNNs with higher robustness than the vanilla ReLU, verifying the flexibility and versatility of the proposed method.
}




\maketitle

%---------------------------------------------------------------------
%---------------------------------------------------------------------
\section{Introduction}\label{sec:intro}

% \felix{need a very intuitive motivation figure to show why spatially adaptive activation handles adversarial scenarios better?}

% \felix{What does it mean to have higher adversarial robustness, when we visualize the data decision boundaries? what do we expect to see?}

% (1) Adversarial poses a challenge for CNNs.

Ever since the identification of the adversarial examples \cite{szegedy2013intriguing} posing severe security threats to deep convolution neural networks (CNNs), studies have been pouring in to improve the adversarial robustness of the CNNs  \cite{papernot2016distillation,buckman2018thermometer,xie2017mitigating,dhillon2018stochastic,liu2018towards,wang2018defensive,bhagoji2018enhancing,guo2017countering,prakash2018deflecting,song2017pixeldefend,samangouei2018defense,liao2018defense}. Among these various methods, adversarial training \cite{goodfellow2014explaining,Kannan2018corr,madry2017towards} is regarded as one of the most effective attempts to improve the adversarial robustness of the neural network. Adversarial training aims at solving a min-max game by training on adversarial examples (on-the-fly) until the model learns to classify them correctly.
%
% Given training data-label pairs $(x,y)\in\mathcal{X}$, loss function $\ell(\cdot)$, CNN $F_\theta$ with network weights $\theta$, and a pre-specified $\epsilon$-ball range where $x$ can perturb, the adversarial training approach \cite{madry2017towards} solves: $\theta^* = \argmin_{\theta} \mathbb{E}_{(x,y)\in\mathcal{X}} [ \max_{\delta\in[-\epsilon,\epsilon]^N}  \ell(x+\delta;y; F_\theta)  ]$.
% \begin{align}\label{eq:minmax}
% \theta^* = \argmin_{\theta} \mathbb{E}_{(x,y)\in\mathcal{X}} \bigg[ \max_{\delta\in[-\epsilon,\epsilon]^N}  \ell(x+\delta;y; F_\theta)   \bigg].
% \end{align}
Specifically, adversarial training is composed of two iterative steps, \ie, an inner $\max$ step that finds the adversarial examples and an outer $\min$ step that carries out network parameters updates. 
Under this paradigm and in this work, we set out to investigate the impacts of basic network components, such as the commonly used ReLU activation, to the adversarial training effectiveness.
%
% Felix added 就是这两个部分“#Introduction第一段第二段之间，最好讲一下activation function尤其是ReLU在CNN中的重要作用，是不是现在所有SOTA的CNNs基本都用ReLU。让读者清楚为什么选择这个ReLU basic component来研究对AT的影响。”
%
Linear rectifier-based activation functions such as ReLU and variants enjoy the following advantages over the traditionally employed `S'-shape activations such as Sigmoid and $\mathrm{tanh}$: \ding{182} less prone to vanishing gradient \cite{glorot2011deep}, \ding{183} more computationally efficient, and \ding{184} better convergence \cite{krizhevsky2012imagenet}.

We argue that the spatially-shared and input-independent activating properties of the ReLU make CNNs under both standard training and adversarial training less robust to white-box adversarial attacks. Such uniformity across input spatial dimensions and different input data may be less ideal in suppressing adversarial patterns, rendering the adversarial training less effective, as we will thoroughly explore in experiments. To address such challenges, we design a novel activation function, \ie, \textsc{Sparta}: spatially-attentional activation for adversarial robustness, by allowing the activation to allocate different amounts of attention across input spatial dimensions, and to be dynamically adapted for each individual input. The flexibility in \textsc{Sparta}, as opposed to the uniformity in ReLU, enables CNNs to achieve higher robustness (\ie, lower error rate on adversarial examples), without hurting accuracy on clean examples, than CNNs based on the SOTA (non-spatially attentional and non-dynamic) activation functions. 

%
%---------------------------
\begin{figure*}
\centering
\includegraphics[width=1.0\textwidth]{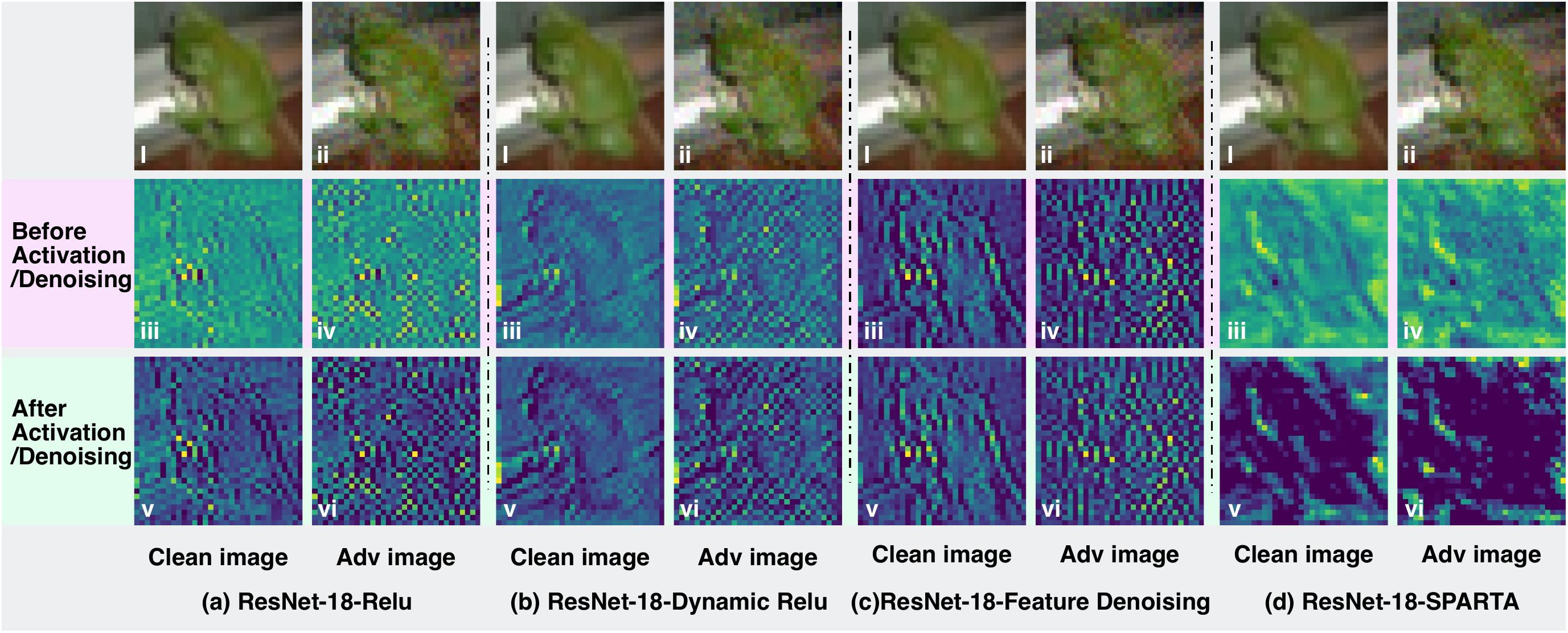}
\caption{
ResNet-18 features of clean images and adversarial images before activation and after activation. We construct ResNet-18s with ReLU, Dynamic ReLU \cite{Chen20ECCV}, feature denoising \cite{Xie_2019_CVPR}, and our \textsc{Sparta} and learn these models via standard training. Then, we feed each ResNet-18 an clean image and an adversarial example crafted from the model itself, respectively.
For each image, we visualize the features before activation (\ie, iii and iv) and after activation (\ie, v and vi) at the last layer of the first group of the ResNet-18-ReLU.
}
\label{fig:cases}
\end{figure*}
%---------------------------
%
%

We further investigate the relationships between our \textsc{Sparta} and the SOTA search-based activation function, \ie, Swish \cite{Ramachandran2017}, and feature denoising method \cite{Xie_2019_CVPR}, providing insights about the advantages of our method. Moreover, comprehensive evaluation demonstrates two important properties of our method: 1) \textit{superior transferability across CNNs}. The adversarially trained activation function for one CNN (\eg, ResNet-18) can be fixed to train another adversarially robust CNN (\eg, ResNet-34), achieving higher robustness than the one using ReLU; 2) \textit{superior transferability across datasets}. The \textsc{Sparta} function trained on one dataset (\eg, CIFAR-10) can be employed to train adversarially robust CNNs on another dataset (\eg, SVHN) and helps achieve higher robustness than CNNs with ReLU. These properties demonstrate the advantage of \textsc{Sparta} in terms of flexibility and versatility. 
% We have posted the preprint version at \cite{guo2021sparta}.

\if 0
Overall, our main contributions can be summarized as follows:
%----------------------------------------
\begin{itemize}
    \item We propose a novel activation function named as \textsc{Sparta} that enables CNNs to achieve higher adversarial robustness and accuracy.
    \item We reveal the relationship between the new activation function and the state-of-the-art activation functions. The insightful analysis shows that our method helps existing activation functions achieve better results.
    \item We show two important properties of our method, \ie, superior transferability across models and datasets, demonstrating the flexibility and versatility of our proposed method.
\end{itemize}
%----------------------------------------
\fi 

% Spatially shared/ input independent activation is (top example in 3.1)

%----------------------------------------------------------
%----------------------------------------------------------
\section{Related Work}\label{sec:related}

{\bf Adversarial training methods.} 
Adversarial training generates adversarial examples on-the-fly for training CNNs \cite{goodfellow2014explaining,Nkland2015ImprovingBB,Kannan2018corr,madry2017towards}.
Athalye \etal\cite{athalye2018obfuscated} demonstrates that projected gradient descent (PGD)-based adversarial training can be considered as the current state-of-the-art defense method.
Tram{\`{e}}r \etal\cite{TramerKPGBM18ICLR} proposes to perform adversarial training with the adversarial examples generated from several pre-trained models.
Then, a series of works are proposed to further enhance the PGD-based adversarial training via \cite{Ciss2017ParsevalNI, Hein2017FormalGO,Xie_2019_CVPR,Farnia2019GeneralizableAT,Li2020EnhancingIA,Goodman2019ImprovingAR}.
% %
In particular, Goodman \etal\cite{Goodman2019ImprovingAR} notes the importance of image-level attention for adversarial training while our work focuses on feature-level attention realized by the newly designed activation function.
% %
% More recently, \cite{Xie_2019_CVPR,Li2020EnhancingIA} proposes to add feature denoisers into the network and further enhances the capability of adversarial training for adversarial robustness.
%
Nevertheless, limited works so far have studied the effects of basic network components such as the ReLU activation to adversarial training. In this work, we discuss the limitations of ReLU for adversarial training (\ie, the spatially-share and input-independent properties) and further propose the spatially-attentional activation function for more higher adversarial robustness. 

{\bf Other Adversarial robustness enhancement methods.} Besides adversarial training, numerous studies have shown to be effective towards enhancing adversarial robustness of CNNs:
(1) The ones that involve non-differentiable operators, intentionally or unintentionally. The introduced non-differentiability and numeric instability lead to incorrect and degenerate gradients such as applying the thermometer encoding \cite{buckman2018thermometer}, performing various image transformations (cropping, bit-depth reduction, \etc) \cite{guo2017countering}, and using local intrinsic dimensionality to characterize adversarial subspaces \cite{ma2018characterizing}. However, they may be circumvented by computing the backward pass using a differentiable approximation of the function \cite{athalye2018obfuscated}.
(2) The ones involve either a randomized network such as \cite{dhillon2018stochastic,liu2018towards,wang2018defensive} or randomly transformed inputs such as \cite{xie2017mitigating,bhagoji2018enhancing,guo2017countering}, which hinder the correct estimation of the true gradient when using a single sample of the randomness. However, they may be countered by computing the gradient correctly over the expected transformation to the input \cite{athalye2018obfuscated}. 
(3) The ones involve input data purification such as high-level representation guided denoiser \cite{liao2018defense}, pixel deflection \cite{prakash2018deflecting}, PixelDefend \cite{song2017pixeldefend}, and Defense-GAN \cite{samangouei2018defense}. However, re-parameterization can greatly diminish these attempts for improving the adversarial robustness of the CNNs \cite{athalye2018obfuscated}. 
(4) Others such as defensive distillation \cite{papernot2016distillation} and adversarially robust architecture \cite{dong2020adversarially}.
%
% Felix add “#Other Adversarial robustness enhancement methods比起AT应该是不属于SOTA methods。是不是应该在这段结尾总结一下，说明它们比起AT的主要不足之处。或者说这些方法可以和AT互补和合并使用？”
%
% Theses methods usually involve 

{\bf Activation functions and attentional modules.} In recent years, quite a few works attempt to study how to improve the ReLU activation function from the viewpoint of enhancing CNNs' accuracy \cite{NairH10ICML,maas2013rectifier,GoodfellowWMCB13ICML,HeZRS15ICCV,ClevertUH16ICLR,Ramachandran2017,Hu2018CVPR,Chen20ECCV,Xie20arXiv}. However, few of them investigate from the viewpoint of the adversarial training. In Sec.~\ref{subsec:relu}, we summarize eight existing representative activation functions and discuss their properties from spatial-wise, dynamic, and attentional properties via Table~\ref{tab:activations}.

%\fei{What is the variable of activation function? The input?}
%---------------------------
\setlength{\intextsep}{4pt}
\setlength{\columnsep}{6pt}
\begin{table}
% % \vspace{-6pt}
% \begin{table}[t]
	\scriptsize
	\caption{Main activation functions for accuracy enhancements and adversarial robustness.}\label{tab:activations}
	\centering
% 	\resizebox{1\linewidth}{!}{
\begin{tabular}{l|l|c|c|c}
	\toprule 
    Designing for	& Activation & Spatial-wise & Dynamic & Attentional\tabularnewline
	\midrule
	\multirow{6}{*}{\makecell{Accuracy \\Enhancement}} & ReLU &  \textcolor{gray}{\ding{55}} & \textcolor{gray}{\ding{55}} & \textcolor{gray}{\ding{55}} \tabularnewline
	& LeakyReLU & \textcolor{gray}{\ding{55}}  & \textcolor{gray}{\ding{55}}  & \textcolor{gray}{\ding{55}} \tabularnewline
	%& Maxout~\cite{GoodfellowWMCB13ICML}   & \textcolor{gray}{\ding{55}}  & \textcolor{gray}{\ding{55}}  & \textcolor{gray}{\ding{55}} \tabularnewline
	& PReLU            & \textcolor{gray}{\ding{55}}  & \textcolor{gray}{\ding{55}}  & \textcolor{gray}{\ding{55}} \tabularnewline
	& ELU          & \textcolor{gray}{\ding{55}}  & \textcolor{gray}{\ding{55}}  & \textcolor{gray}{\ding{55}} \tabularnewline
	& GELU          & \textcolor{gray}{\ding{55}}  & \textcolor{gray}{\ding{55}}  & \textcolor{gray}{\ding{55}} \tabularnewline
	& Swish         & \textcolor{gray}{\ding{55}}  & \textcolor{gray}{\ding{55}}  & \textcolor{gray}{\ding{55}} \tabularnewline
	%& SE~\cite{Hu2018CVPR}                 & \textcolor{gray}{\ding{55}}  & \textbf{\ding{51}}  & \textcolor{gray}{\ding{55}} \tabularnewline
	& Dynamic ReLU      & \textbf{\ding{51}}  & \textbf{\ding{51}}  & \textcolor{gray}{\ding{55}} \tabularnewline
	\midrule 
	\multirow{2}{*}{\makecell{Adversarial\\ Robustness}} & Swish & \textcolor{gray}{\ding{55}}  & \textcolor{gray}{\ding{55}}  & \textcolor{gray}{\ding{55}} \tabularnewline
	& \textsc{Sparta} (Ours) & \textbf{\ding{51}}  & \textbf{\ding{51}}  & \textbf{\ding{51}} \tabularnewline
	\bottomrule 
\end{tabular}
% }
% \end{table}
% % \vspace{-15pt}
\end{table}
%---------------------------
%

% % \vspace{-7pt}

\section{Existing ReLU Activations and Challenges}\label{subsec:relu}

Given an input tensor $\mathbf{X}$, the widely used activation function, \eg, ReLU, can be represented as
%
%---------------------------
\begin{align}\label{eq:relu}
\mathbf{Y}_p=\max(\mathbf{X}_p,0),~\forall p\in \mathcal{P},
\end{align}
%---------------------------
%
where $\mathbf{X}_p$ is the $p$-th element in $\mathbf{X}$ and $\mathcal{P}$ denotes the set of all element positions of $\mathbf{X}$. 
The corresponding derivative of this function \wrt the input $\mathbf{X}$ is 
%
%---------------------------
\begin{align}\label{eq:relubp}
\frac{d \mathbf{Y}_p}{d \mathbf{X}_p} =
\left\{\begin{matrix}
1,~&\text{if}~\mathbf{X}_p\geq0, \\
0,~&\text{if}~\mathbf{X}_p< 0,
\end{matrix}
\right.
~~\forall p\in \mathcal{P}.
\end{align}
%---------------------------
%
We argue that \textit{such a unified activation across all elements of input tensor during the both forward and backward processes makes the adversarial training less effective.}

To make it clear, we briefly introduce the adversarial attack. Given an example, an adversarial attack feeds it to a targeted CNN and calculate the loss via a loss function $\mathcal{L}$. Then, the attack conducts back-propagation and gets the gradient of loss \wrt to the input example. After that, the gradient is used to produce the adversarial perturbations. By conducting these steps iteratively, we obtain the final adversarial perturbations that are added to the original example and the adversarial example is generated.
When we feed the adversarial example to the targeted CNN, the corresponding adversarial perturbations can be propagated to deep layers of the CNN and lead to feature noise and misclassification \cite{Xie_2019_CVPR}. 
From the viewpoint of the activation function in the CNN (\ie, \reqref{eq:relu} and \reqref{eq:relubp}), the white-box adversarial attack can be easily achieved due to: \ding{182}  for the forward process, both clean and perturbed elements in $\mathbf{X}$ are equally activated, making the adversarial noise easily propagate to the deep layers, thus affecting the prediction results directly. 
To validate this, we construct ResNet-18 with ReLU and learn it via standard training. Then, we feed the ResNet-18 an clean image and an adversarial example crafted from the model itself, respectively.
For each image, we can visualize the features before and after activation at the last layer of the first group of the ResNet-18-ReLU. 
As the results of ResNet-18-ReLU shown in \figref{fig:cases}, the features of the adversarial image before activation (\ie, \figref{fig:cases} (a)-iv) present obvious noise patterns compared with the features of the clean image (\ie, \figref{fig:cases} (a)-iii). When the noisy features are further passed through the ReLU, the noisy patterns remain in the activated features (\ie, \figref{fig:cases} (a)-vi) and are very different from the features of the clean image (\ie, \figref{fig:cases} (a)-v).
\ding{183} during the back-propagation of the white-box attack, the gradients of all elements in $\mathbf{X}$ evenly pass through the activation function for generating the adversarial perturbations, making the white-box attack re-search optimized solution easily. Specifically, when considering the back-propagation on the activation function, we have the gradient of loss \wrt the output of an activation (\ie, $\nabla_{\mathbf{Y}_p}\mathcal{L}$) that is back-propagated from the last layer of the CNN. Then, we can calculate the gradient of loss \wrt the input of the activation (\ie, $\nabla_{\mathbf{X}_p}\mathcal{L}$) by $\nabla_{\mathbf{X}_p}\mathcal{L}=\nabla_{\mathbf{Y}_p}\mathcal{L}\cdot \frac{d \mathbf{Y}_p}{d \mathbf{X}_p}$.
When we use the ReLU activation function, we have $\frac{d \mathbf{Y}_p}{d \mathbf{X}_p}=1$ for any $\mathbf{X}_p\ge 0$ according to the derivative function in \reqref{eq:relubp}. If the $\mathbf{X}_p$ is the element that should be corrupted for effective white-box attack, the gradient of $\mathbf{X}_p$ can be back-propagated to the previous layers without any attenuation via $\nabla_{\mathbf{X}_p}\mathcal{L}=\nabla_{\mathbf{Y}_p}\mathcal{L}$. As a result, the adversarial perturbations can be optimized effectively and the CNN is easily fooled.
We will validate these in \secref{subsec:analysis}.
% As shown in Fig.~\ref{fig:pgdloss}, during the targeted adversarial attack, the loss is easily minimized when we use the vanilla ReLU.

To overcome the above limitations, we take the following two factors into consideration to design the novel activation function: \ding{182} a spatial-wise and attentional activation should be developed, which makes different elements in $\mathbf{X}$ have different activation conditions and semantic-related elements be preserved while perturbations being suppressed.
For example, the elements corrupted by adversarial noise should be suppressed during the activating while the clean ones should be preserved. \ding{183} The activation should be dynamic, that is, it could be tuned to adapt to different inputs. Actually, the first factor indicates that the activation should have the spatial-wise and attentional properties, where the semantic and clean elements should be highlighted while the corrupted ones should be suppressed. The second factor indicates that the activated value of each element should consider the whole input.

Although some recent attempts have been made to explore how to improve the ReLU \cite{NairH10ICML} from the angle of accuracy enhancement, including LeakyReLU \cite{maas2013rectifier}, PReLU \cite{HeZRS15ICCV}, ELU \cite{ClevertUH16ICLR}, GELU \cite{hendrycks2016gaussian}, Swish \cite{Ramachandran2017}, Dynamic ReLU \cite{Chen20ECCV}, and Swish \cite{Xie20arXiv}, none of them could perfectly fit the above two key factors. 
We summarize their basic information in terms of the spatial-wise, attentional and dynamic properties in Table~\ref{tab:activations}. Among these improved ReLU variants, LeakyReLU and exponential linear unit (ELU) extend the activation range to negative values while all input elements share the same activation condition, which cannot be tuned according to inputs.
PReLU adds extra learnable parameters to the basic ReLU, which are trainable but fixed for different inputs after training.
%
%SE \felix{what is SE?} further lets the activation rely on the input and realize dynamic activation shared by all elements.
%
Dynamic ReLU is a spatial-wise and dynamic activation function where each input element has an exclusive activation function represented by several linear functions whose slope and bias parameters are dynamically predicted by a network.
Nevertheless, dynamic ReLU is specifically designed for accuracy enhancement, which lacks generality and does not consider the attentional requirement of adversarially robustness, failing to suppress adversarial corrupted elements.
For example, we present the results of normally trained ResNet-18-Dynamic ReLU shown in \figref{fig:cases} and observe: the features of the adversarial image before activation present obvious noise patterns (\ie, \figref{fig:cases} (b)-iv) and the noisy patterns remain in the activated features (\ie, \figref{fig:cases} (b)-vi). As a result, the activated features of the adversarial image vary greatly from those of the clean image.
We further discuss the quantitative and qualitative results in Sec.~\ref{subsec:cmp_acts}.
In addition to above activations, MaxOut \cite{GoodfellowWMCB13ICML} and squeeze-and-excitation networks (SE) \cite{Hu2018CVPR} can be also used to realize activations as introduced in \cite{Chen20ECCV}.
MaxOut has learnable parameters for the ReLU and can be offline trained but the parameters cannot change according to different inputs.
SE lets the activation rely on the input and realizes dynamic activation that however is shared by all elements.

%----------------------------------------------------------
%----------------------------------------------------------
\section{Methodology}\label{sec:method}

% In this section, we introduce our spatially attentional activation function and investigate its effects to the adversarial robustness under standard training and adversarial training, so that to answer a key question: whether or not the spatial-wise, dynamic, and attentional activation can benefit CNN's adversarial robustness?

\subsection{Spatially Attentional Activation Function} \label{subsec:form_arch}
%
% \subsubsection{Formulation}\label{subsec:formulation}
%
{\bf Formulation.} To address the robustness challenges, we propose the \textsc{Sparta}. Given an input tensor $\mathbf{X}$, we have
%
%---------------------------
\begin{align}\label{eq:sparta}
\mathbf{Y}_p=\max(\mathbf{X}_p,0)\cdot\phi_\theta(\mathbf{X})[p],~\forall p\in \mathcal{P},
\end{align}
%---------------------------
%
where $\phi_\theta(\cdot)$ is a sub-network with $\theta$ as the parameters. The sub-network takes all elements of $\mathbf{X}$ as inputs, predicts a new tensor that has the same size with $\mathbf{X}$, and assigns a weight for each element of $\mathbf{X}$. Hence, $\phi_\theta(\mathbf{X})[p]$ denotes the $p$-th element of $\phi_\theta(\mathbf{X})$ and we have $\{0\leq \phi_\theta(\mathbf{X})[p]\leq 1 \| \forall p, p\in \mathcal{P}\}$. Then, we get the derivative of Eq.~\eqref{eq:sparta} \wrt the input $\mathbf{X}$
%
%---------------------------
\begin{align}\label{eq:spartabp}
\frac{d \mathbf{Y}_p}{d \mathbf{X}_p} =
\left\{\begin{matrix}
\phi_\theta(\mathbf{X})[p]+\mathbf{X}_p\frac{\partial \phi_\theta(\mathbf{X})}{\partial \mathbf{X}_p},&\text{if}~\mathbf{X}_p\geq0, \\
0,&\text{if}~\mathbf{X}_p< 0,
\end{matrix}
\right.
~~\forall p\in \mathcal{P}
\end{align}
%---------------------------
%
When comparing \reqref{eq:sparta} and \reqref{eq:spartabp} with the forward and backward processes of ReLU (\ie, \reqref{eq:relu} and \reqref{eq:relubp}), we notice that: \ding{182} for the forward process, each activated element is further processed by a scalar estimated from $\phi_\theta(\mathbf{X})$ that considers the whole input. Intuitively, the pre-trained $\phi_\theta(\cdot)$ decides whether the $p$-th element of $\mathbf{X}$ should be suppressed according to the understanding of the whole input. \ding{183} In terms of the backward process, in contrast to Eq.~\eqref{eq:relubp}, the activated elements' gradients are not propagated to the earlier layers directly but determined by $\phi_\theta(\mathbf{X})$ and  $\mathbf{X}_p\frac{\partial \phi_\theta(\mathbf{X})}{\partial \mathbf{X}_p}$. 
When $\phi_\theta(\cdot)$ is a deep neural network with the Sigmoid function as the last layer for activation, the gradient of its input $\frac{\partial \phi_\theta(\mathbf{X})}{\partial \mathbf{X}_p}$ tends to be very small \cite{Glorot2010AISTATS}. Then, we can say that $\frac{d \mathbf{Y}_p}{d \mathbf{X}_p}$ mainly relies on $\phi_\theta(\mathbf{X})[p]$, meaning that the white-box attack based on back-propagation is affected by the $\phi_\theta(\mathbf{X})[p]$. 
For example, if $\mathbf{X}_p$ is the element that should be adversarially corrupted for an effective attack and we have $\frac{d \mathbf{Y}_p}{d \mathbf{X}_p}\approx\phi_\theta(\mathbf{X})[p]<1$, the gradient of the loss \wrt the $\mathbf{X}_p$ is suppressed and becomes $\nabla_{\mathbf{X}_p}\mathcal{L}\approx\nabla_{\mathbf{Y}_p}\mathcal{L}\cdot\frac{d \mathbf{Y}_p}{d \mathbf{X}_p}<\nabla_{\mathbf{Y}_p}\mathcal{L}$. As a result, the white-box attack would be harder to be optimized due to the less effective back-propagated gradients.
We will validate the two concerns under both standard training and adversarial training in Sec.~\ref{subsec:analysis}. Note that, Eq.~\eqref{eq:spartabp} does not harm the CNN's accuracy under standard training and can even be helpful to achieve lower error rate. As shown in Table~\ref{tab:relu_vs_sparta}, ResNet-18s with our activation function (\ie, \textsc{Sparta}-w/o-DPNet and \textsc{Sparta} that will be introduced in Sec.~\ref{subsec:form_arch}) achieve lower top-1 error rate than the network using ReLU under both standard and adversarial training.

% %-------------------------------------------------------------------------%
% \begin{table}[t]
% % \vspace{-5pt}
% % \begin{table}[t]
% 	\scriptsize
% 	\caption{Comparing ResNet-18s equipped with ReLU, \textsc{Sparta}-w/o-DPNet, and \textsc{Sparta}, respectively.} \label{tab:relu_vs_sparta}
% 	\centering
%     \resizebox{1\linewidth}{!}{
%     \begin{tabular}{l|l|c|c|c|c}
%     \toprule
%     \multirow{2}{*}{} & \multirow{2}{*}{ResNet-18} & \multicolumn{3}{c|}{Top-1 error on Adv. Images} & \multirow{2}{*}{\makecell{Top-1 err. on\\Clean Img}}\tabularnewline
%     %
%      &  & PGD-10 & PGD-30 & PGD-50 & \tabularnewline
%     \midrule
%     \multirow{3}{*}{\makecell{Adv. Train. \\on PGD-10}} & ReLU & 31.54\%  & 68.93\% & 75.64\% & 15.66\%\tabularnewline
%     %
%      & \makecell[l]{\textsc{Sparta}\\-w/o-DPNet} & 29.43\% & 66.13\% & 72.35\% & 15.44\%\tabularnewline
%     %
%      & \textsc{Sparta} & 29.31\% & 65.81\% & 72.55\% & 15.48\%\tabularnewline
%     \midrule
%     \multirow{3}{*}{Std. Train.} & ReLU & 100.0\% & 100.0\% & 100.0\% & 7.71\%\tabularnewline
%     %
%      & \makecell[l]{\textsc{Sparta}\\-w/o-DPNet} & 99.94\% & 100.0\% & 100.0\% & 6.98\%\tabularnewline
%     %
%      & \textsc{Sparta} & 99.85\% & 100.0\% & 100.0\% & 6.90\%\tabularnewline
%     \bottomrule
%     \end{tabular}
%     }
% % \end{table}
% % \vspace{-15pt}
% \end{table}
% %-------------------------------------------------------------------------%
% %

%-------------------------------------------------------------------------%
\setlength{\intextsep}{0.1pt}
\setlength{\columnsep}{0.1pt}
\begin{table}[t]
% \vspace{10pt}
% \begin{table}[t]
	\scriptsize
	\caption{ Comparing ResNet-18s equipped with ReLU, \textsc{Sparta}-w/o-DPNet, \textsc{Sparta}-on-ActFeat, and \textsc{Sparta} on clean CIFAR-10, corresponding PGD-10, PGD-30, and PGD-50 adversarial examples under adversarial and standard training, respectively.}
	\label{tab:relu_vs_sparta}
	\centering
    % \resizebox{1.0\linewidth}{!}{
    \begin{tabular}{l|l|c|c|c|c}
    \toprule
    \multirow{2}{*}{} & \multirow{2}{*}{ResNet-18} & \multicolumn{3}{c|}{Top-1 error on Adv. Images} & \multirow{2}{*}{\makecell{Top-1 err. on\\Clean Img}}\tabularnewline
     &  & PGD-10 & PGD-30 & PGD-50 & \tabularnewline
    \midrule
    \multirow{4}{*}{\makecell{Adv. Train. \\on PGD-10}} & ReLU & 31.54\%  & 68.93\% & 75.64\% & 15.66\%\tabularnewline
    & \makecell[l]{\textsc{Sparta}\\-on-ActFeat} & 29.80\% & 66.45\% & 73.56\% & 15.31\% \tabularnewline
     & \makecell[l]{\textsc{Sparta}\\-w/o-DPNet} & 29.43\% & 66.13\% & 72.35\% & 15.44\%\tabularnewline
     & \textsc{Sparta} & \topone{29.31\%} & \topone{65.81\%} & \topone{72.55\%} & 15.48\%\tabularnewline
    \midrule
    \multirow{4}{*}{Std. Train.} & ReLU & 100.0\% & 100.0\% & 100.0\% & 7.71\%\tabularnewline
    & \makecell[l]{\textsc{Sparta}\\-on-ActFeat} & 99.87\% & 100\% & 100\% & 7.29\% \tabularnewline
    & \makecell[l]{\textsc{Sparta}\\-w/o-DPNet} & 99.94\% & 100.0\% & 100.0\% & 6.98\%\tabularnewline
     & \textsc{Sparta} & 99.85\% & 100.0\% & 100.0\% & 6.90\%\tabularnewline
    \bottomrule
    \end{tabular}
    % }
% \end{table}
% \vspace{-15pt}
\end{table}
%-------------------------------------------------------------------------%
%

% % \vspace{-7pt}
% \subsubsection{Architecture of $\phi_\theta(\cdot)$}\label{subsec:archs} 
%
%
%---------------------------
\begin{figure}[t]
\centering
\includegraphics[width=0.55\columnwidth]{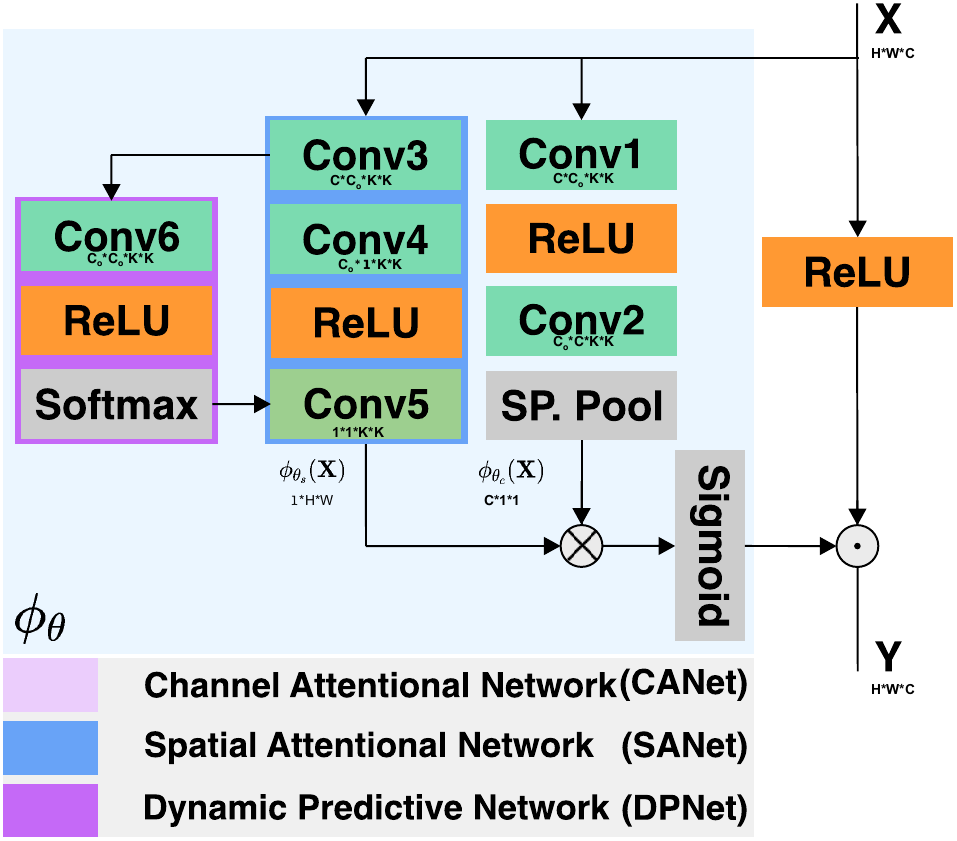}
\caption{Proposed \textit{dynamic spatial-channel-attentional network} ($\phi_\theta(\cdot)$) containing 3 sub-networks, \ie, channel attentional net ($\phi_{\theta_c}(\cdot)$), spatial attentional net ($\phi_{\theta_s}(\cdot)$), and dynamic predictive net ($\phi_{\theta_d}(\cdot)$), where $\theta=\{\theta_s,\theta_c,\theta_d\}$. The Sp. Pool performs spatial pooling and generate channel attentional vector.}
\label{fig:archs}
% \vspace{-18pt}
\end{figure}
%---------------------------
%

%
%---------------------------
\begin{figure*}
\centering
\includegraphics[width=1.0\textwidth]{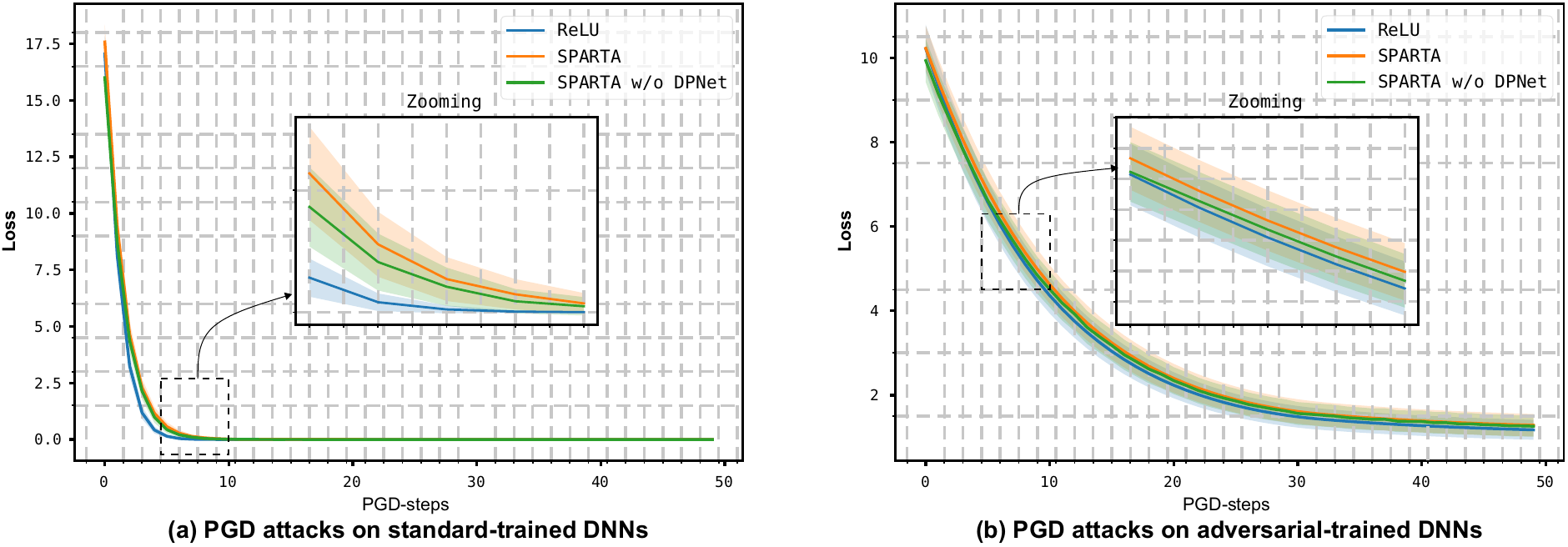}
\caption{Comparing loss values of ResNet-18 with ReLU, \textsc{Sparta}, and \textsc{Sparta}-w/o-DPNet during the PGD attack under standard training (a) and adversarial training (b).
}
\label{fig:pgdloss}%
\end{figure*}
%---------------------------
%

{\bf Architecture of $\phi_\theta(\cdot)$.} A simple architecture for $\phi_\theta(\cdot)$ can be a CNN that takes $\mathbf{X}$ as the input and outputs a tensor having the same size to decide the activation weights of each element.
However, such an architecture requires a large number of parameters, making the whole network difficult to train.
Moreover, since an activation can be deployed at different locations of a CNN (\ie, from the shallow layers to deep ones), the inputs $\mathbf{X}$ would be diverse (\eg, the $\mathbf{X}$ from shallow layers mainly contain spatial details while the deep ones focus on semantic information). Hence, a dynamic architecture that tunes the attentional network according to the input is highly desired.
To this end, we propose the \textit{dynamic spatial-channel-attentional network (DSCANet)} as $\phi_\theta(\cdot)$
%
%---------------------------
\begin{align}\label{eq:sp_ch_atten}
\phi_\theta(\mathbf{X}) = \text{Sigmoid}(\phi_{\theta_s}(\mathbf{X})\otimes\phi_{\theta_c}(\mathbf{X})),
\end{align}
%---------------------------
%
where `$\otimes$' is the outer production, and $\phi_{\theta_s}(\cdot)$ and $\phi_{\theta_c}(\cdot)$ denote the spatial-attentional network (SANet) and channel-attentional network (CANet), respectively.
%
% \footnote{\scriptsize For brevity, \emph{spatially attentional} and \emph{channel-wise attentional} networks are sometimes paraphrased as \emph{spatial-attentional} and \emph{channel-attentional} networks.} 
%
When we have $\mathbf{X}\in \mathds{R}^{H\times W\times C}$, $\phi_{\theta_s}(\mathbf{X})\in\mathds{R}^{H\times W}$ is the spatial attentional map across all channels and $\phi_{\theta_c}(\mathbf{X})\in\mathds{R}^{1\times 1\times C}$ is the channel attentional vector across all spatial positions. 
Moreover, inspired by the dynamic convolution \cite{chen2020dynamic}, we construct a dynamic predictive network (DPNet) to predict the parameters of the last layer of SANet according to the input. As a result, we realize the desired dynamic property of activation. As shown in \figref{fig:archs}, the whole architecture contains three sub-networks where the sizes of convolution layers and tensors are shown at the bottom, and Conv5's parameters are estimated from DPNet.
% the Conv1, Conv2, Conv3, and Conv6 are with the size of `$C\times C\times 3\times 3$', Conv4 is with the size of `$C\times 1\times 3\times 3$', and Conv5 is with the size of `$1\times 1\times 3\times 3$'. 
%
In practice, we set the input channel numbers of Conv3 and Conv1 as $C_\text{o}=\min(256,C)$ and the kernel size of convolution layers (\ie, $K$ shown in the Fig.~\ref{fig:archs}) as $1\times 1$ to avoid heavy costs. 
We detail the complexity analysis and compare costs on parameters with other activations in \secref{subsec:cmp_acts}. 
We will further discuss the influence of different architectures in the \secref{subsec:analysis}.
% %
% %---------------------------
% \begin{align}\label{eq:sp_ch_atten}
% \theta_{s} = [\theta_{s1},\theta_{s2}],~\theta_{s2} = \phi_{\theta_d}(\mathbf{X}).
% \end{align}
% %---------------------------
% %

\subsection{Analysis of \textsc{Sparta}}\label{subsec:analysis}

We aim to analyze \textsc{Sparta} by comparing with the basic ReLU in Sec.~\ref{subsec:relu}, and answer the following questions: Does the \textsc{Sparta} help achieve higher adversarial robustness under \emph{standard training} and \emph{adversarial training}, respectively? Do the advantages stem from the spatial-wise, dynamic, and attentional architectures? 
Moreover, we discuss how to perform replacement with \textsc{Sparta} in a CNN.

\subsubsection{Setup}\label{subsec:anaylysis_setup}
For a comprehensive analysis of the proposed activation function, we use ResNet-18 \cite{He2016CVPR} as the backbone network and modify it by replacing the last ReLU layers of the four groups in ResNet-18 with our \textsc{Sparta}, respectively. %
We further discuss the influence of replacement strategies in Sec~\ref{subsubsec:sparta_loc}.
Then, we conduct the image classification task on CIFAR-10 dataset, comparing the top-1 error rate of the raw ResNet18 and the modified one under both standard training and adversarial training.
For adversarial training, we follow the setups in \cite{Xie_2019_CVPR} and perform the targeted projected gradient descent (PGD) attack \cite{MadryMSTV2018ICLR} to generate adversarial examples with 
% the step size of $1.0$, and 
the maximum perturbation of $16.0$. The targeted class is selected uniformly at random. The same setup is
also used in the testing. We implement three PGD attacks according to the iteration number of $10$, $30$, and $50$ and denote them as PGD-10, PGD-30, and PGD-50, respectively.
Note that, in all sub-sequence experiments, the top-1 error rate on adversarial (adv.) images means that we first generate adversarial examples by using one attack method to attack the evaluated CNN and calculate the error rate on these adversarial images.
The sub-network and CNN are jointly trained, where we set the learning rate to be 0.1 with the $10\times$ attenuation at the $30$th and $60$th epochs, and the weight decay is set to be $1\text{e}{-4}$. In Sec.~\ref{subsec:trans_dnns}, we also show that pre-trained \textsc{Sparta} on one model and dataset can be fixed and benefit adversarial training of another model.

\subsubsection{Dynamic and attentional activation benefits adversarial robustness} \label{subsec:analysis_reluvssparta}

Table~\ref{tab:relu_vs_sparta} summarizes the top-1 error rates of three versions of ResNet-18 on the adversarial and clean images of CIFAR-10 under both adversarial and standard training, from which we have the following observations and conclusions: 
\ding{182} Compared with ReLU, \textsc{Sparta} does let the CNN achieve much better adversarial robustness (\ie, lower top-1 error rate on adversarial examples from white-box PGD attacks, \eg, 75.64\% vs. 72.55\% on PGD-50) under adversarial training while further improving the accuracy on clean images (\eg, 15.66\% vs. 15.48\%), concluding that \textit{\textsc{Sparta} improves adversarial robustness without the sacrifice of classification accuracy for clean images}.
\ding{183} In terms of the standard training, \textsc{Sparta} leads to lower top-1 error rate (\ie, 100\% for ReLU vs. 99.85\% for \textsc{Sparta}) under the PGD-10 while achieving much higher accuracy (\ie, lower error rate on clean images), demonstrating that \textit{\textsc{Sparta} does not rely on adversarial training and still benefits to both adversarial robustness and accuracy under standard training.}
\ding{184} Compared with \textsc{Sparta}-w/o-DPNet where the DPNet in $\phi_\theta(\cdot)$ is removed, \textsc{Sparta} achieves lower top-1 errors under all PGD attacks in the cases of adversarial and standard training, confirming that \textit{the dynamic activation function via the proposed DPNet helps the CNN achieve higher adversarial robustness.}
\ding{185} When further comparing ReLU with \textsc{Sparta}-w/o-DPNet, we see that ResNet-18 with \textsc{Sparta}-w/o-DPNet has lower top-1 error rates on most of the PGD attacks and clean images. Under the standard training, \textsc{Sparta}-w/o-DPNet always improves the CNN with lower error rates on clean images. These results demonstrate \textit{attentional activation introduced by the spatial attentional network and channel attentional network does enhance the CNNs' adversarial robustness and accuracy and also benefits the standard training for higher accuracy}.

To better understand the above results, we conduct an experiment to compare the loss values of pre-trained CNNs during PGD attacks, to validate if \textsc{Sparta} makes the adversarial attack harder as explained in Sec.~\ref{subsec:form_arch}. Specifically, we perform PGD-50 on 20\% examples of CIFAR-10 testing dataset and collect the loss values during the optimization process. Then, we calculate the mean and standard deviation of loss values at each iteration step across all examples, and draw three plots of the ResNet-18s with ReLU, \textsc{Sparta}-w/o-DPNet, and \textsc{Sparta}, respectively.
As shown in Fig.~\ref{fig:pgdloss}, we see that the loss values of CNNs based on \textsc{Sparta} and \textsc{Sparta}-w/o-DPNet are always larger than that of ReLU-based CNNs along the iteration. It demonstrates that \textit{the proposed attentional and dynamic activation function does make the optimization of adversarial attack harder for both adversarial and standard trained CNNs.}

%
%---------------------------
\begin{figure*}
\centering
\includegraphics[width=1.0\textwidth]{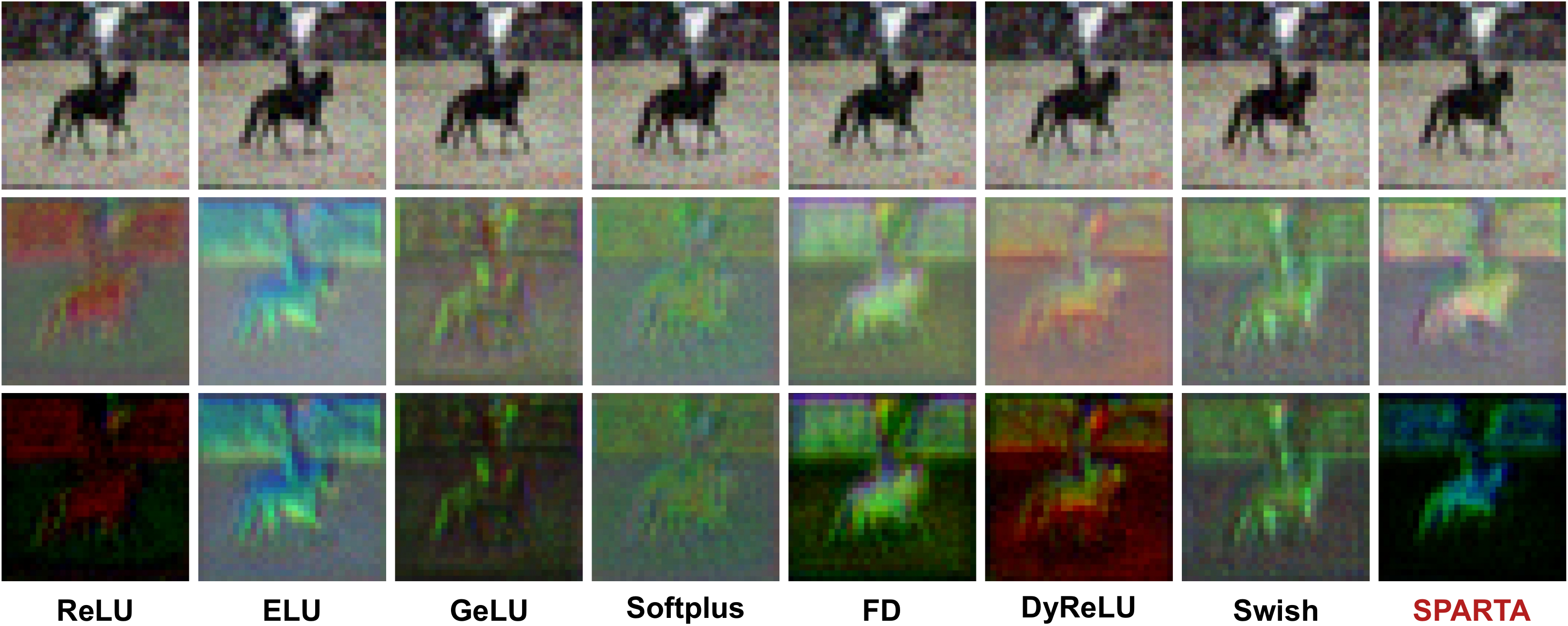}
\caption{
Features before and after activation of ResNet-18s with eight activation functions. The first row presents the input adversarial images that are crafted from the ResNet-18s with the activation functions indicated at the bottom row. The second and third rows present the features before and after activation at the last layer of the first group of each ResNet-18, respectively. Note that, all ResNet-18s are learned via adversarial training.
}
\label{fig:vis}
\end{figure*}
%---------------------------
%

% \subsubsection{Level of Adaptivity of the Activations vs. network adversarial robustness, with and without AdvTraining}

% % \vspace{-7pt}
\subsubsection{Spatial-wise activation benefits adversarial robustness}
In addition to the attentional and dynamic properties, we further analyze the importance of spatial-wise activation. To this end, we set spatial-neighboring elements of $\mathbf{X}$ sharing the same attentional scores and control the neighboring size to study the influence of spatial-wise activation. We reformulate Eq.~\eqref{eq:sparta} to represent the above process as
%
%---------------------------
\begin{align}\label{eq:sparta_sw}
\mathbf{Y}_p=\max(\mathbf{X}_p,0)\cdot\phi_\theta(\mathbf{X})[\lfloor\frac{p}{N}\rfloor],~\forall p\in \mathcal{P},
\end{align}
%---------------------------
%
where we set the output size of $\phi_\theta(\mathbf{X})$ to be $\frac{H}{N}\times \frac{W}{N}\times C$ and $N$ controls the neighboring size. For example, when we set $N=2$, every four elements in $\mathbf{X}$ share the same attentional scores; when we have $N=1$, Eq.~\eqref{eq:sparta_sw} becomes Eq.~\eqref{eq:sparta}, \ie, each element has its exclusive attentional score.
%
%
%-------------------------------------------------------------------------%
\begin{table}[t]
% \vspace{-5pt}
\scriptsize
\caption{Comparing ResNet-18s equipped with \textsc{Sparta} according to different spatial-wise setups defined by Eq.~\eqref{eq:sparta_sw}, respectively. The best results are highlighted.} \label{tab:sparta_sw}
\centering
\begin{tabular}{c|c|c|c|c}
\toprule
\multirow{2}{*}{\makecell{ResNet-18 \\with \textsc{Sparta}}} & \multicolumn{3}{c|}{Top-1 error on Adv. Images} & \multirow{2}{*}{\makecell{Top-1 err on\\Clean Img}} \tabularnewline
& PGD-10 & PGD-30 & PGD-50 &  \tabularnewline
\midrule
$N=1$ & \topone{29.73\%} & \topone{65.93\%} & \topone{72.98\%} & 15.25\% \tabularnewline
$N=2$ & 30.11\% & 66.27\% & 73.40\% & 14.98\% \tabularnewline
$N=4$ & 30.17\% & 66.58\% & 73.45\% & 15.07\% \tabularnewline
\bottomrule
\end{tabular}
% \vspace{-15pt}
\end{table}
%-------------------------------------------------------------------------%
%
To analyze the effects of different $N$, we modify ResNet-18 by replacing the fourth block's ReLU with \textsc{Sparta} and obtain three CNNs by setting $N=1,2,4$. Then, we perform the adversarial training and evaluate the robustness and accuracy, respectively.
%
% \fei{the adversarial examples are generated on old model or retrained model?}. 
%
We present the results in Table~\ref{tab:sparta_sw} with the following observations: \ding{182} for all three PGD attacks, the top-1 error rate on adversarial images increases as the $N$ becomes larger (\ie, more elements share the same attentional score), indicating that \textit{spatial-wise activation benefits the adversarial robustness.} \ding{183} In terms of the results on clean images, the CNN with $N=1$ has higher top-1 error rate than the ones with $N=2$ and $N=4$, which indicates the spatial-wise activation could reduce the accuracy to some extent. Even though, \textsc{Sparta} with $N=1$ still enables the CNN to achieve lower error rate than ReLU. 

%
%-----------------------------------------------------------
\begin{table}[t]
    % % \vspace{-5pt}
	\scriptsize
% 	% % \vspace{1mm}
	\caption{Comparing ResNet-18s with \textsc{Sparta} employed at different depths. The best results are highlighted.} \label{tab:sparta_loc}
	\centering
    \begin{tabular}{c|c|c|c|c}
    \toprule
    \multirow{2}{*}{\makecell{ResNet-18: replacing \\ReLU with \textsc{Sparta} at}} & \multicolumn{3}{c|}{Top-1 error on Adv. Images} & \multirow{2}{*}{\makecell{Top-1 err on\\Clean Img}}\tabularnewline
     & PGD-10  & PGD-30 & PGD-50 & \tabularnewline
    \midrule
    G$_1$.B$_2$ & 32.33\% & 67.23\% & 74.38\% & 17.41\%\tabularnewline
    G$_2$.B$_2$ & 31.61\% & 67.76\% & 75.50\% & 15.68\%\tabularnewline
    G$_3$.B$_2$ & 31.15\% & 68.12\% & 75.01\% & 15.77\%\tabularnewline
    G$_4$.B$_2$ & \topone{29.73\%} & \topone{65.93\%} & \topone{72.98\%} & 15.25\%\tabularnewline
    \midrule
    G$_{\{1,2,3,4\}}$.B$_{\{1,2\}}$ & 31.32\% & 65.83\% & \topone{72.28\%} & 16.70\%\tabularnewline
     G$_{\{1,2,3,4\}}$.B$_{2}$ & \topone{29.31\%} & \topone{65.81\%} & 72.55\% & 15.48\%\tabularnewline
    \bottomrule
    \end{tabular}
    % % \vspace{-15pt}
\end{table}
%-----------------------------------------------------------

\subsubsection{Effects of \textsc{\textbf{Sparta}}'s Locations in CNNs} \label{subsubsec:sparta_loc}
Since \textsc{Sparta} contains extra parameters for the three sub-networks in Fig.~\ref{fig:archs}, it is ideal to perform as fewer ReLU replacements as possible to avoid heavy costs. To this end, we study the influence of replacement positions based on the widely used ResNet and take the representative ResNet-18 as a representative case to study.
ResNet-18 contains four groups and each group has two blocks. 
We focus on replacing the last ReLU layer of each block and set the following strategies: \textit{First}, we replace the last ReLU layer of the second block of each group with our \textsc{Sparta} and obtain four CNNs denoted as ResNet-18-G$_{i}$.B$_2$ where $i$ denotes the $i$th group of ResNet-18 and B$_2$ represents the second block. This setup helps explore the influence of \textsc{Sparta} at different depths of a CNN.
\textit{Second}, we perform the replacements on all groups simultaneously and study whether more substitutions lead to better adversarial robustness. In particular, we consider two versions denoted as 1) G$_{\{1,2,3,4\}}$.B$_{\{1,2\}}$ and 2) G$_{\{1,2,3,4\}}$.B$_{2}$, respectively. The first one replaces the last ReLU layers of the two blocks of all groups, while the second one only conducts replacement on the second block of all groups.

\subsection{Relationship to Existing Methods and Beyond}\label{subsec:relationship}
{\bf Relationship to Swish and beyond.} More recently, \cite{Xie20arXiv} identifies the importance of the smooth activation function for adversarial training and shows the search-based activation function, \ie, Swish \cite{Ramachandran2017,Stefan2018sigmoid}, which achieves the state-of-the-art adversarial robustness and can be represented as
%---------------------------
\begin{align}\label{eq:swish}
\mathbf{Y}_p=\mathbf{X}_p\cdot\text{Sigmoid}(\mathbf{X}_p),~\forall p\in \mathcal{P}.
\end{align}
%---------------------------
Meanwhile, we can reformulate the \textsc{Sparta} by combining
Eq.~\eqref{eq:sparta} and \eqref{eq:sp_ch_atten} and have
%---------------------------
% {\footnotesize
\begin{align}\label{eq:sparta_new}
&\mathbf{Y}_p=\max(\mathbf{X}_p,0)\cdot\text{Sigmoid}((\phi_{\theta}(\mathbf{X}))[p]),\\
& \phi_{\theta}(\mathbf{X})[p]=(\phi_{\theta_s}(\mathbf{X})\otimes\phi_{\theta_c}(\mathbf{X}))[p],~\forall p\in \mathcal{P}.
\end{align}
% }
%---------------------------
Comparing Eq.~\eqref{eq:swish} with Eq.~\eqref{eq:sparta_new}, we can see that Swish is very similar to our \textsc{Sparta}, but having two major differences: \ding{182} the first term $\mathbf{X}_p$ in the right part of Eq.~\eqref{eq:swish} is further processed via
ReLU in Eq.~\eqref{eq:sparta_new}. \ding{183} In terms of the variable in $\text{Sigmoid}(\cdot)$, Eq.~\eqref{eq:swish} uses $\mathbf{X}_s$ itself while Eq.~\eqref{eq:sparta_new} adopts the spatial and channel attentions that consider all elements in $\mathbf{X}$. We have demonstrated the advantages of the spatial and channel attentions in Sec.~\ref{subsec:analysis_reluvssparta}. 
Here, we further study the influence of the first difference and show that the \textit{adversarial robustness of Swish can be further enhanced by simply adding the ReLU to Eq.~\eqref{eq:swish}}. As shown in Table~\ref{tab:swish_sparta}, when equipping Swish with ReLU (\ie, Swish-ReLU), the top-1 error rates on all PGD attacks decrease.

%-----------------------------------------------------------
\begin{table}[t]
% \vspace{-5pt}
\scriptsize
\caption{Comparing ResNet-18s with Swish, Swish-ReLU, and \textsc{Sparta} under 3 PGD attacks. The best results are highlighted.} \label{tab:swish_sparta}
\centering
\begin{tabular}{l|c|c|c}
\toprule 
\multirow{2}{*}{ResNet-18 with} & \multicolumn{3}{c}{Top-1 error on Adv. Images}\tabularnewline
 & \multicolumn{1}{c}{PGD-10} & \multicolumn{1}{c}{PGD-30} & PGD-50\tabularnewline
\midrule
Swish & 30.38\% & 68.65 \% & 76.21\%\tabularnewline
Swish-ReLU & 30.10\% & 67.70\% & 75.07\%\tabularnewline
\textsc{Sparta} & \topone{29.31\%} & \topone{65.81\%} & \topone{72.55\%} \tabularnewline
\bottomrule
\end{tabular}
% \vspace{-15pt}
\end{table}
%-----------------------------------------------------------

%-----------------------------------------------------------
\begin{table*}[t]
% \vspace{-5pt}
\scriptsize
\caption{Comparing \textsc{Sparta} with ReLU, ELU, GELU, feature denoising operation (FD), Dynamic ReLU (DyReLU), Softplus, and Swish by equipping them to ResNet-18 and ResNet-34 for adversarial training and standard training on CIFAR-10 dataset. The best results are highlighted.} \label{tab:cmp_cifar}
\centering
% \resizebox{1.0\linewidth}{!}{
\begin{tabular}{l|c|c|c|c|c}
\toprule% \hline 
% \multirow{4}{*}{} & \multicolumn{5}{c}{ResNet-18} \tabularnewline
% \cline{2-11} \cline{3-11} \cline{4-11} \cline{5-11} \cline{6-11} \cline{7-11} \cline{8-11} \cline{9-11} \cline{10-11} \cline{11-11} 
\multirow{4}{*}{ResNet-18} & \multicolumn{4}{c|}{Adv. Training} & Std. Training \tabularnewline
% \cline{2-11} \cline{3-11} \cline{4-11} \cline{5-11} \cline{6-11} \cline{7-11} \cline{8-11} \cline{9-11} \cline{10-11} \cline{11-11} 
 & \multicolumn{3}{c|}{Error on Adv. Imgs} & \multirow{2}{*}{\makecell{Error on\\ Clean Imgs}} & \multirow{2}{*}{\makecell{Error on\\ Clean Imgs}} \tabularnewline
% \cline{2-4} \cline{3-4} \cline{4-4} \cline{7-9} \cline{8-9} \cline{9-9} 
 & PGD-10 & PGD-30 & PGD-50 &  & \tabularnewline
\midrule % \hline 
ReLU & 31.54\% & 68.93\% & 75.64\% & 15.66\% & 7.71\% \tabularnewline
% \hline 
ELU & 31.44\% & 67.57\% & \toptwo{72.98\%} & 17.08\% & 7.99\% \tabularnewline
% \hline 
GELU & \toptwo{30.27\%} & 68.95\% & 75.30\% & 14.55\% & 6.85\% \tabularnewline
% \hline 
Softplus & 33.08\% & 69.65\% & 74.92\% & 16.81\% & 9.35\%      \tabularnewline
% \hline 
FD & 31.81\% & \toptwo{67.03\%} & 73.52\% & 16.81\% & 7.99\% \tabularnewline
% \hline 
DyReLU & 32.57\% & 68.54\% & 75.08\% & 16.97\% & 7.40\% \tabularnewline
% \hline 
Swish & 30.38\% & 68.65\% & 76.21\% & 14.26\% & 7.00\% \tabularnewline
% \hline 
\textsc{Sparta} & \topone{29.31\%} & \topone{65.81\%} & \topone{72.55\%} & 15.48\% & 6.90\% \tabularnewline
%-------------------------------------------
\midrule % \hline 
% \multirow{4}{*}{} & \multicolumn{5}{c}{ResNet-34}\tabularnewline
%
\multirow{4}{*}{ResNet-34} & \multicolumn{4}{c|}{Adv. Training} & Std. Training \tabularnewline
& \multicolumn{3}{c|}{Error on Adv. Imgs} & \multirow{2}{*}{\makecell{Error on\\ Clean Imgs}} & \multirow{2}{*}{\makecell{Error on\\ Clean Imgs}} \tabularnewline
                 & PGD-10 & PGD-30 & PGD-50 &  & \tabularnewline
\midrule
ReLU             & 31.00\% & 67.88\% & 75.15\% & 17.18 \% & 7.36\% \tabularnewline
ELU              & 31.31\% & 67.67\% & 74.02\% & 17.16\% & 7.88\% \tabularnewline
GELU             & \toptwo{29.66\%} & 65.95\% & \toptwo{73.68\%} & \toptwo{15.03\%} & 6.52\% \tabularnewline
Softplus         & 33.91\% & 68.99\% & 74.57\% & 18.41\% & 8.11\% \tabularnewline
FD               & 30.71\% & \toptwo{65.72\%} & 73.90\% & 17.49\% & 8.05\% \tabularnewline
DyReLU           & 31.56\% & 68.40\% & 75.84\% & 17.63\% & 8.78\% \tabularnewline
Swish            & 32.20\% & 67.51\% & 74.42\% & 16.74\% & 7.28\%  \tabularnewline
\textsc{Sparta}  & \topone{29.17\%} & \topone{64.13\%} & \topone{72.91\%} & \topone{14.47\%} & 6.78\% \tabularnewline
\bottomrule% \hline 
\end{tabular}
% }
% \vspace{-15pt}
\end{table*}
%---------------------------------------------------------------------

{\bf Relationship to feature denoiser.}
Xie \etal \cite{Xie_2019_CVPR} improves the adversarial robustness of CNNs by adding extra blocks for feature denoising, inspired by the fact that the pixel-level adversarial noise poses large perturbations to the deep features and leads to noisy activation overwhelming the true ones, resulting in erroneous predictions.
\textsc{Sparta} can also be regarded as a denoising block and has the capability of feature denoising, since the perturbed elements in $\mathbf{X}$ are selectively activated or suppressed according to the predictive results of $\phi_\theta(\mathbf{X})$. 
Compared with the method of \cite{Xie_2019_CVPR}, \textsc{Sparta} has the following difference and advantages: \ding{182} from the viewpoint of denoising, \textsc{Sparta} uses multiplication for denoising and dynamically tunes the parameters via DPNet according to the inputs while the feature denoising method adopts the addition with fixed denoising operations. The higher level of flexibility of \textsc{Sparta} helps CNNs achieve much better adversarial robustness. 
As shown in Fig.~\ref{fig:cases}, in terms of the ResNet-18-\textsc{Sparta} under standard training, we see obvious noise patterns of the adversarial image before activation (\ie, \figref{fig:cases} (d)-iv), which are suppressed by \textsc{Sparta} (See \figref{fig:cases} (d)-vi). As a result, the features after activation becomes similar with the clean ones (Compare \figref{fig:cases} (d)-vi and v). In contrast, the feature denoising is less effective for feature noise removal and the noise patterns of the adversarial image before and after denoising are similar.
We conduct more detailed qualitative and quantitative analysis in \secref{subsec:cmp_acts}.
\ding{183} \textsc{Sparta} is a new activation function that can directly replace existing ReLUs in a CNN without changing its original architecture. 
On the other hand, the feature denoising method needs to add new blocks to existing CNNs, requiring extra adaption costs.

{\bf Relationship to existing attentional module.} 
There are some works that also use spatial and channel attentions to improve the performance of other tasks. In particular, Chen \etal \cite{chen2017sca} employ spatial-wise and channel-wise attentive features as encoders that are fed to an LSTM decoder for description words generation. Our \textsc{Sparta} is totally different from such an attentional module \cite{chen2017sca}: \textit{First}, the main motivation and objectives are different. \textsc{Sparta} is to replace the traditional activation function and improve the adversarial robustness of backbone networks via adversarial training. In contrast, the attention module \cite{chen2017sca} is to enhance the features extracted from an fixed backbone network to achieve accurate descriptions.
\textit{Second}, the main solutions for attentions are different. The work \cite{chen2017sca} uses activated features (\ie, $\max(\mathbf{X},0)$) to predict attentions while our method employs the raw features (\ie, $\mathbf{X}$). Here, we argue that such a difference is important for enhancing the adversarial robustness since using the activated features to predict attentions neglects the unactivated elements.
To validate this, we consider a variant of \textsc{Sparta} that predicts weights via the activated feature maps (\ie, $\max(\mathbf{X},0)$) and reformulate \reqref{eq:sparta} in the main manuscript as
%
%---------------------------
\begin{align}\label{eq:sparta_plus}
\mathbf{Y}_p=\max(\mathbf{X}_p,0)\cdot\phi_\theta(\max(\mathbf{X},0))[p],~\forall p\in \mathcal{P}.
\end{align}
%---------------------------
%
Compared with \reqref{eq:sparta}, \reqref{eq:sparta_plus} feeds the activated feature maps $\max(\mathbf{X},0)$ instead of $\mathbf{X}$ to the $\phi_\theta(\cdot)$ predicting the attentional weights. We name this version as \textsc{Sparta}-on-Act.Feat. and equip it to ResNet-18. Then, we evaluate the results on PGD-10/30/50 attacks under adversarial training and standard training. As the results shown in \tableref{tab:relu_vs_sparta}, \textsc{Sparta}-on-Act.Feat. gets higher top-1 errors than our final \textsc{Sparta} on all scenarios including clean images and adversarial examples of PGD-10/30/50 under adversarial training and standard training. The results demonstrate that using the raw features $\mathbf{X}$ to predict spatial and channel attentions help achieve higher accuracy and adversarial robustness than using activated features.
\textit{Third}, we design a novel \textit{dynamic spatial-channel-attentional network} to estimate the attentions. For example, in \secref{subsec:analysis_reluvssparta}, we have demonstrated that the dynamic predictive network is able to enhance the robustness of backbone network significantly.

% %
% %-----------------------------------------------------------
% \begin{table}
% % \vspace{-5pt}
% 	\scriptsize
% 	\caption{Comparing \textsc{Sparta} with 6 activations on SVHN datasets.
% 	The \textcolor{red}{best} and \textcolor{purple}{second} results are highlighted.
% 	} \label{tab:cmp_svhn}
% 	\centering
% \begin{tabular}{l|c|c|c}
% \toprule
% \multirow{2}{*}{ResNet-18 with} & \multicolumn{3}{c}{Top-1 error on Adv. Images} \tabularnewline %& \multirow{2}{*}{\makecell{Top-1 error \\on Clean Images}}
% % \cline{2-4} \cline{3-4} \cline{4-4} 
%  & PGD-10 & PGD-30 & PGD-50 \tabularnewline
% \midrule 
% ReLU & 18.27\% & 63.71\% & 71.42\% \tabularnewline %& 6.45\%
% % \hline 
% ELU & 18.74\% & 63.38\% & 70.72\% \tabularnewline %& 6.45\%
% % \hline 
% GELU & 16.76\% & 61.21\% & 69.32\% \tabularnewline  %& 6.15\%
% %
% FD & 16.45\% & 61.85\% & 69.45\% \tabularnewline  %& 6.15\%
% % \hline 
% DyReLU & 17.11\% & 63.50\% & 70.71\% \tabularnewline %& 7.09\%
% % \hline 
% Swish & \toptwo{16.41\%} & \toptwo{61.00\%} & \toptwo{68.97\%}  \tabularnewline %& 6.90\%
% % \hline 
% \textsc{Sparta} & \topone{15.30\%} & \topone{59.29\%} & \topone{67.29\%} \tabularnewline %& 7.05\%
% \bottomrule
% \end{tabular}
% % \vspace{-15pt}
% \end{table}
% %-----------------------------------------------------------
% %

%
%-----------------------------------------------------------
\setlength{\intextsep}{2pt}
\setlength{\columnsep}{3pt}
\begin{table*}[t]
	\scriptsize
	\caption{Comparing with ReLU, ELU, GELU, feature denoising operation (FD), Dynamic ReLU (DyReLU), Softplus, and Swish by equipping them to ResNet-18 for adversarial training on Tiny-ImageNet and SVHN datasets. The best results are highlighted.} \label{tab:cmp_svhn_tinyimage}
	\centering
\begin{tabular}{l|c|c|c|c}
\toprule% \hline 
% \multirow{4}{*}{} & \multicolumn{4}{c|}{ResNet-18 on Tiny-ImageNet} & \multicolumn{4}{c}{ResNet-18 on SVHN}\tabularnewline
% \cline{2-11} \cline{3-11} \cline{4-11} \cline{5-11} \cline{6-11} \cline{7-11} \cline{8-11} \cline{9-11} \cline{10-11} \cline{11-11}
%
\multirow{3}{*}{\makecell{ResNet-18 on \\Tiny-ImageNet}} & \multicolumn{3}{c|}{Top-1 Err. on Adv. Imgs} & \multirow{2}{*}{\makecell{Top-1 Err. on\\ Clean Imgs}} \tabularnewline
% \cline{2-4} \cline{3-4} \cline{4-4} \cline{7-9} \cline{8-9} \cline{9-9} 
        & PGD-10 & PGD-30 & PGD-50 &  \tabularnewline
\midrule % \hline 
ReLU    & \toptwo{67.38\%} & 84.11\% & 86.46\% & 54.32\% \tabularnewline
% \hline 
ELU     & 67.98\% & 85.40\% & 86.87\% & 55.06\% \tabularnewline
% \hline 
GELU    & 67.86\% & 84.12\% & \toptwo{85.60\%} & 54.90\% \tabularnewline
% \hline 
Softplus & 68.60\% & 85.08\% & 86.75\% & 54.95\% \tabularnewline
% \hline 
FD      & 70.21\% & 84.94\% & 86.35\%& 57.63\% \tabularnewline
% \hline 
DyReLU  & 67.82\% & \toptwo{84.06\%} & 85.90\% & 55.61\% \tabularnewline
% \hline 
Swish  & 67.95\% & 85.19\% & 86.15\% & 54.89\% \tabularnewline
% \hline 
% \textsc{Sparta}-v1  & 67.68\% & \topone{83.25\%} & 85.71\% & 54.72\% & 68.05\% & \toptwo{82.63\%} & \topone{83.74\%} & 54.67\% \tabularnewline
\textsc{Sparta}  & \topone{67.21\%} & \topone{83.62\%} & \topone{84.70\%} & 54.89\% \tabularnewline
\midrule
%-----------------------------------------
%
\multirow{3}{*}{\makecell{ResNet-18 \\on SVHN}} & \multicolumn{3}{c|}{Top-1 Err. on Adv. Imgs} & \multirow{2}{*}{\makecell{Top-1 Err. on\\ Clean Imgs}} \tabularnewline
& PGD-10 & PGD-30 & PGD-50 & \tabularnewline
\midrule
ReLU    & 18.27\% & 63.71\% & 71.42\% & 6.44\% \tabularnewline
% \hline 
ELU     & 18.74\% & 63.38\% & 70.72\% & 6.70\% \tabularnewline
% \hline 
GELU    & 16.76\% & 61.21\% & 69.32\% & 6.98\% \tabularnewline
% \hline 
Softplus & 18.62\% & 63.96\% & 70.58\% & 6.82\% \tabularnewline
% \hline 
FD      & 16.45\% & 61.85\% & 69.45\% & 6.15\%  \tabularnewline
% \hline 
DyReLU  & 17.11\% & 63.50\% & 70.71\% & 7.09\% \tabularnewline
% \hline 
Swish   & \toptwo{16.41\%} & \toptwo{61.00\%} & \toptwo{68.97\%}  & 6.90\%\tabularnewline
% \hline 
\textsc{Sparta}  & \topone{15.30\%} & \topone{59.29\%} & \topone{67.29\%} & 6.54\% \tabularnewline
\bottomrule% \hline 
\end{tabular}
% \vspace{-6pt}
\end{table*}
%---------------------------------------------------------------------

%----------------------------------------------------------
%----------------------------------------------------------
\section{Experiments}\label{sec:exp}

\subsection{Setup}
\label{subsec:exp_setup}
Following the setup in Sec.~\ref{subsec:anaylysis_setup}, we further consider ResNet-18 and ResNet-34 \cite{He2016CVPR} as the backbones and evaluate on CIFAR-10 \cite{krizhevsky2009learning}, Tiny-ImageNet, and SVHN \cite{Netzer2011NIPS}, mainly investigating five questions: 
How is the performance of \textsc{Sparta} compared with state-of-the-art activation functions, including ReLU \cite{NairH10ICML}, ELU \cite{ClevertUH16ICLR}, GELU \cite{hendrycks2016gaussian}, Softplus \cite{nair2010rectified}, Swish \cite{Ramachandran2017,Xie20arXiv}, Dynamic ReLU (DyReLU) \cite{Chen20ECCV}, and the feature denoising method (FD) \cite{Xie_2019_CVPR}?
How is the transferability of CNNs with different activation functions across different attacks?
Can \textsc{Sparta} be shared across CNNs, 
can \textsc{Sparta} be shared across datasets, and
can \textsc{Sparta} be generalized to more advanced AT methods?
Note that, we select ELU, GELU, and Softplus as activation baselines since their capability for adversarial robustness enhancement has been studied in \cite{Xie20arXiv}, \cite{pang2020bag}, and \cite{gowal2020uncovering}.

\subsection{Comparison with State-of-the-art Activations} 
\label{subsec:cmp_acts}

\textbf{Adversarial robustness results.} We compare our \textsc{Sparta} with six SOTA activations and the feature denoising method through ResNet-18 and ResNet-34 architectures. Note that, we implement all baseline activation functions according to their public released codes. As the results on CIFAR-10 shown in Table~\ref{tab:cmp_cifar}, we have the following observations: \ding{182} Under adversarial training, CNNs with \textsc{Sparta} achieve the lowest top-1 error on all three levels of PGD attack as well as lower top-1 error than CNNs with ReLU, demonstrating that \textit{the proposed activation does help CNNs realize better adversarial robustness without sacrificing the accuracy.} 
\ding{183} Under standard training, CNNs with \textsc{Sparta} have the second best accuracy (\ie, second lowest top-1 error on clean images), indicating that \textit{the proposed activation architecture not only benefits to adversarial training for better robustness but also helps achieve higher accuracy.}
Then, we further conduct the comparison on SVHN and Tiny-ImageNet datasets and present the results in Table~\ref{tab:cmp_svhn_tinyimage}. Similar with the results on CIFAR-10, our \textsc{Sparta} has lower top-1 errors than all other baseline methods under PGD-10, 30, and 50 attacks with similar accuracy with ReLU on the clean images, which further demonstrates the advantages of our method against adversarial attacks. Note that, we also evaluate all activations on other architectures like VGG-16 and VGG-19 and observe that \textsc{Sparta} also achieves the best adversarial robustness.

\textbf{Visualization results.} We visualize and compare the features before and after activation of the ResNet-18s based on eight activation functions. 
Note that, the visualized features are three-channel and extracted from the last layer of the first group of ResNet-18s and all CNNs are trained via adversarial training.
As shown in \figref{fig:vis}, we have the following observations: \ding{182} Baseline activation functions fail to suppress the introduced noise effectively. For example, ReLU makes the main object information invisible in the background. \ding{183} Our \textsc{Sparta} can not only suppress the adversarial noise but also highlight the object regions due to the design of the spatial-wise attention.

% In addition to the error rate comparison,  we further compare \textsc{Sparta} and six baseline methods about their model size in \tableref{tab:cmp_cifar}. Compared with baseline methods, we see that \textsc{Sparta}-v2 with similar model size achieves lower error rate under PGD-30 and PGD-50.

%-------------------------------------------------------------------------%
\setlength{\intextsep}{2pt}
\setlength{\columnsep}{1pt}
\begin{table}[t]
% \vspace{-5pt}
% \begin{table}[t]
	\scriptsize
	\caption{Computing cost of ResNet-18 under different activations on CIFAR-10 dataset. Testing time denotes the averaging time cost per image on the testing dataset of CIFAR-10. Training time is the averaging time cost for per example on the training dataset of CIFAR-10, which includes the adversarial attack cost.} 
	\label{tab:complexity_cmp}
	\centering
    % \resizebox{1\linewidth}{!}{
    \begin{tabular}{l|c|c|c}
    \toprule
    ResNet-18 & Param.~Num.  & \makecell{Test Time \\/per~image~(ms)} & \makecell{Train Time \\/per~image~(s)} \tabularnewline
    \midrule
    \makecell[l]{ReLU}             & 11.17~M & 2.87 & 1.57 \tabularnewline
    \makecell[l]{ELU}              & 11.17~M & 2.85 & 1.67 \tabularnewline
    \makecell[l]{GELU}             & 11.17~M & 3.28 & 2.58 \tabularnewline
    \makecell[l]{SoftPlus}         & 11.17~M & 2.89 & 2.22 \tabularnewline
    \makecell[l]{FD}               & 11.87~M & 4.06 & 4.02 \tabularnewline
    \makecell[l]{DyReLU}           & 11.79~M & 3.61 & 3.40 \tabularnewline
    \makecell[l]{Swish}           & 11.17~M & 3.08 & 2.62 \tabularnewline
    \makecell[l]{\textsc{Sparta}}  & 11.99~M & 3.79 & 3.68 \tabularnewline
    \bottomrule
    \end{tabular}
    % }
% \end{table}
% \vspace{-15pt}
\end{table}
%-------------------------------------------------------------------------%

%
\textbf{Complexity analysis.}
As detailed in the \secref{subsec:form_arch} and \figref{fig:archs}, the model size of \textsc{Sparta} is determined by the input/output channel numbers (\ie, $C$ and $C_\text{o}$) and the kernel size of convolution layers for the three sub-networks (\ie, $K$ shown in the Fig.~\ref{fig:archs}).
To avoid heavy cost, we set $K=1$ and the channel-related parameter, \ie,  $C_\text{o}=\min(256,C)$.
To make it clear, in \tableref{tab:complexity_cmp}, we report the model sizes, training, and testing time costs of ResNet-18s with eight different activation functions including ReLU, ELU, GELU, Softplus, Feature denoising method, dynamic ReLU (DyReLU), Swish, and \textsc{Sparta}. 
We see that ResNet-18 with $\textsc{Sparta}$ has slightly larger parameter number than other versions. For example, ResNet-18 with \textsc{Sparta} has 11.99 million parameters while the models with dynamic ReLU (DyReLU) and feature denoising (FD) methods have 11.87 and 11.79 million parameters, respectively. 
In terms of the testing time costs per example, ResNet-18 with $\textsc{Sparta}$ runs at 3.79 millisecond on average, which is faster than the model with feature denoising (FD) and has similar cost with the model with DyReLU and SmReLU. We have similar results on training time costs.

\subsection{Transferability across Attacks}
\label{subsec:trans_attacks}

%
%---------------------------
\begin{figure}[t]
\centering
\includegraphics[width=0.65\columnwidth]{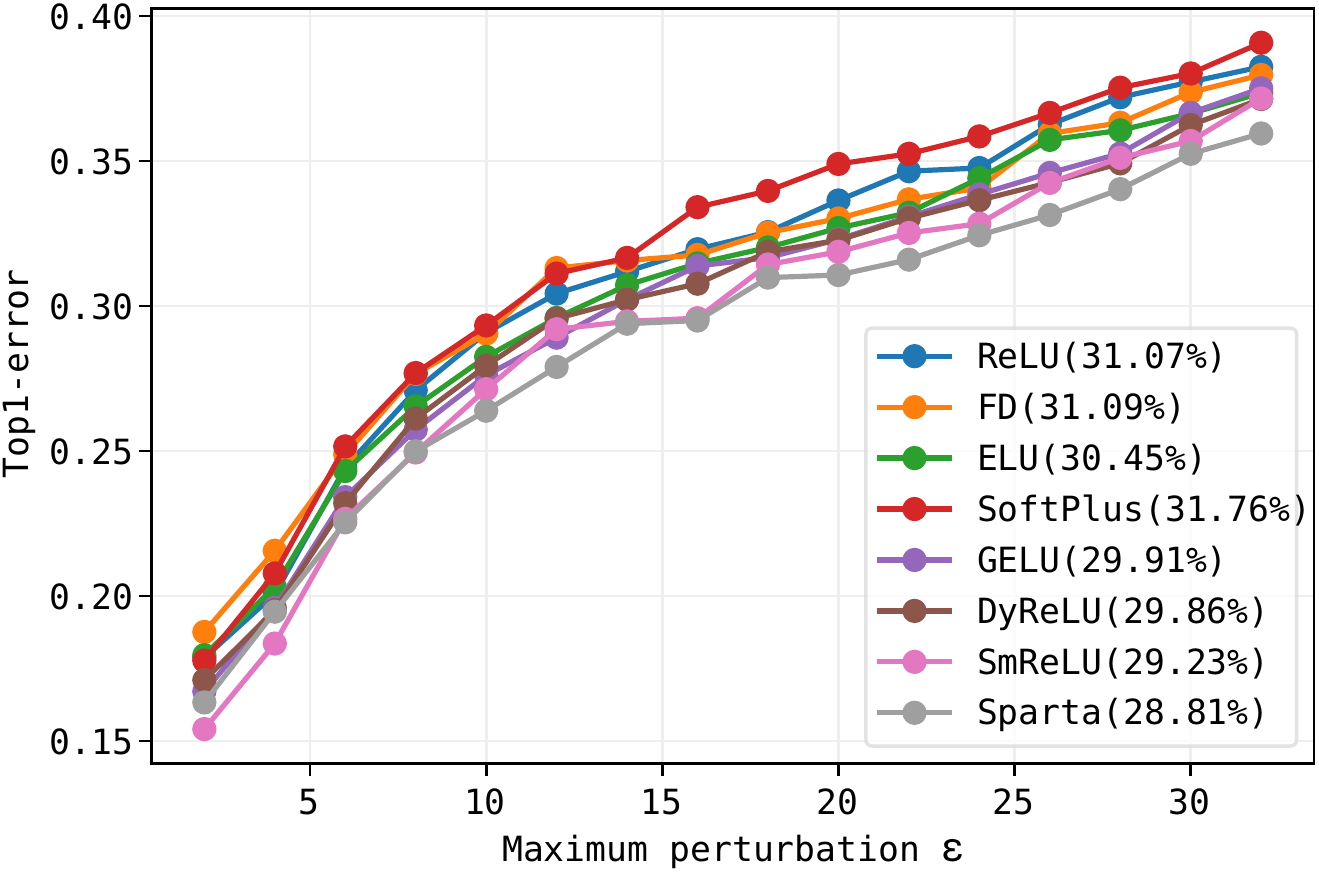}
\caption{Top-1 errors of the adversarially trained ResNet-18 with eight activate functions under different attack sizes (\ie, $\epsilon=[\frac{2}{255},\frac{4}{255},\frac{6}{255},\ldots,\frac{32}{255}]$). We report the averaging top-1 errors after the legend name.}
\label{fig:attacksizes}
% \vspace{-18pt}
\end{figure}
%---------------------------
%

In this subsection, we study the transferability of adversarially trained ResNet18 with different activation functions against different attacks and maximum perturbations.

{\bf Transferability across attacks with different maximum perturbations.} We evaluate the adversarially trained ResNet-18s with different activation functions on CIFAR-10 under sixteen PGD-10 attacks with sixteen maximum perturbations ranging from $\frac{2}{255}$ to $\frac{32}{255}$ with the interval of $\frac{2}{255}$. As shown in \figref{fig:attacksizes}, we see that the DNN with \textsc{Sparta} achieves the lowest average top-1 error (\ie, $28.81\%$) among all compared activations. Specifically, \textsc{Sparta} gets the lowest top-1 errors with the maximum perturbations from $\frac{6}{255}$ to $\frac{32}{255}$ and the second lowest top-1 errors under maximum perturbations of $\frac{2}{255}$ and $\frac{4}{255}$. These results demonstrate the advantages of \textsc{Sparta} on adversarial robustness enhancement in particular for large maximum perturbations. 

{\bf Transferability across different attack methods.}
We train ResNet-18s with different activation functions adversarially under the PGD-10 attack, and evaluate the top-1 errors under other attacks including ADef \cite{alaifari2018adef}, C\&Ws with $L_\infty$ norm and $L_2$ norm \cite{carlini2017towards}, and AutoAttack \cite{croce2020reliable,croce2021robustbench}. Note that, in contrast to previous examples, all models are trained with PyTorch to fit the implementation of other attacks \cite{kim2020torchattacks}.
We see that the ResNet-18 with \textsc{Sparta} achieves the lowest top-1 errors when we attack models with different attacks, which demonstrates advantages of \textsc{Sparta} over other activation methods on the generalization capability.

%
%-----------------------------------------------------------
\setlength{\intextsep}{2pt}
\setlength{\columnsep}{3pt}
\begin{table}[t]
	\scriptsize
	\caption{Comparing the transferability across attacks of ResNet-18s with ReLU, ELU, GELU, Softplus, feature denoising operation (FD), Dynamic ReLU (DyReLU), and Smooth ReLU (SmReLU) under AT on CIFAR-10. All models are trained under PGD-10.} \label{tab:cmp_trans_attacks}
	\centering
\begin{tabular}{l|c|c|c|c|c}
\toprule
\multirow{2}{*}{\makecell[l]{ResNet-18s}} & \multicolumn{5}{c}{Top-1 error on CIFAR-10 dataset under different attacks} \tabularnewline
& PGD & ADef & AutoAttack & CW$_{L_\infty}$ & CW$_{L_2}$ \tabularnewline
\midrule 
ReLU        & 33.77 & \toptwo{54.11} & 97.74 & \toptwo{64.96} & 24.68 \tabularnewline
ELU         & 32.49  & 55.33 & 98.33 & 66.22 & 26.37 \tabularnewline
GELU        & \toptwo{31.56} & 55.07 & 98.12 & 80.72 & \toptwo{22.87} \tabularnewline
Softplus    & 37.19 & 56.73 & 98.44 & 70.56 & 28.77 \tabularnewline
FD          & 36.91 & 55.77 & 97.67 & 72.07 & 23.19 \tabularnewline
DyReLU      & 40.90 & 54.33 & \toptwo{97.49} & 73.46 & 27.38 \tabularnewline
SmReLU      & 31.64 & 55.33 & 98.11 & 65.93 & 23.61 \tabularnewline
\textsc{Sparta} & \topone{31.47} & \topone{53.73} & \topone{97.38} & \topone{62.79} & \topone{21.11} \tabularnewline
\bottomrule
\end{tabular}
\vspace{-6pt}
\end{table}
%---------------------------------------------------------------------

\subsection{Transferability across CNNs} \label{subsec:trans_dnns}
We study the transferability of \textsc{Sparta} across CNNs, \ie, we regard the pre-trained \textsc{Sparta} borrowed from one CNN as the activation function for another CNN and see if it achieves better adversarial robustness. We take ResNet-18 and ResNet-34 as the backbones and conduct the following steps based on CIFAR-10 dataset: \textit{First}, we adversarially train a CNN (\eg, ResNet-34) equipped with \textsc{Sparta} and obtain the pre-trained \textsc{Sparta} at different blocks.
We denote the pre-trained activations as \textsc{Sparta}$_\text{Res34}$.  \textit{Second}, we equip another CNN (\eg, ResNet-18) with \textsc{Sparta}$_\text{Res34}$ and perform the adversarial training without updating the parameters of \textsc{Sparta}$_\text{Res34}$. 
Similarly, we can also train ResNet-34 by using the pre-trained \textsc{Sparta} from ResNet-18 (\ie, \textsc{Sparta}$_\text{Res18}$).
As shown in Table~\ref{tab:sparta_trans_dnns}, we see that: \ding{182} CNNs (\ie, ResNet-18 and ResNet-34) with the transferred \textsc{Sparta} achieve lower top-1 error rate than CNNs using ReLU under all three attacks (\ie, PGD-10, 30, 50) and the clean images. Such results demonstrate that \textit{pre-trained \textsc{Sparta} has the transferability to some extent and can help other CNNs achieve better adversarial robustness and accuracy than the ones using ReLU.}
\ding{183} Compared with the CNNs with standard \textsc{Sparta} (\eg, ResNet-34 with \textsc{Sparta}$_{\text{Res34}}$) whose parameters are jointly updated during adversarial training, the CNNs with transferred \textsc{Sparta} (\eg, ResNet-34 with \textsc{Sparta}$_{\text{Res18}}$) get worse adversarial robustness (\ie, higher error rate under the three attacks). For example, ResNet-34 with \textsc{Sparta}$_{\text{Res34}}$ obtains much lower top-1 error rates than ResNet-34 with \textsc{Sparta}$_{\text{Res18}}$ under all 3 PGD attacks and also has 
slightly lower error on clean images, \ie, $14.47$\% vs. $14.78$\%.

\begin{table}[t]
    % % \vspace{-5pt}
	\scriptsize
	\caption{Comparing the transferred \textsc{Sparta} with the standard \textsc{Sparta} and ReLU. Better results are highlighted.} \label{tab:sparta_trans_dnns}
	\centering
    \begin{tabular}{l|l|c|c|c|c}
    \toprule
    \multirow{2}{*}{Backbone} & \multirow{2}{*}{Activations} & \multicolumn{3}{c|}{Top-1 error on Adv. Images} & \multirow{2}{*}{\makecell{Top-1 err on\\Clean Img}}\tabularnewline
     &  & PGD-10 & PGD-30 & PGD-50 & \tabularnewline
    \midrule
    \multirow{3}{*}{ResNet-18} & ReLU & 31.54\% & 68.93\% & 75.64\% & 15.66\%\tabularnewline
     & \makecell{\textsc{Sparta}$_{\text{Res34}}$} & \toptwo{31.06\%} & \toptwo{68.11\%} & \toptwo{75.03\%} & 15.17\%\tabularnewline
     & \makecell{\textsc{Sparta}$_{\text{Res18}}$} & \topone{29.31\%} & \topone{65.81\%} & \topone{72.55\%} & 15.48\%\tabularnewline
    \midrule 
    \multirow{3}{*}{ResNet-34} & ReLU & 31.00\% & 67.88\% & 75.15\% & 17.18\%\tabularnewline
     & \makecell{\textsc{Sparta}$_{\text{Res18}}$} & \toptwo{29.47\%} & \toptwo{65.50\%} & \toptwo{73.21\%} & 14.78\%\tabularnewline
     & \makecell{\textsc{Sparta}$_{\text{Res34}}$} & \topone{29.17\%} & \topone{64.13\%} & \topone{72.91\%} & 14.47\%\tabularnewline
    \bottomrule
    \end{tabular}
    % % \vspace{-15pt}
\end{table}
%-----------------------------------------------------------
%

% % \vspace{-7pt}
\subsection{Transferability across Datasets}
We further study the transferability of \textsc{Sparta} across datasets.
Specifically, we adversarially train a ResNet-18 with \textsc{Sparta} on CIFAR-10 and obtain the pre-trained \textsc{Sparta} denoted as \textsc{Sparta}$_\text{CIFAR}$. 
Then, we regard \textsc{Sparta}$_\text{CIFAR}$ as the activation function for another random initialized ResNet-18 and adversarially train it on SVHN to see whether the ResNet-18 with \textsc{Sparta}$_\text{CIFAR}$ achieves better adversarial robustness or higher accuracy than the one with ReLU and standard \textsc{Sparta} that are jointly trained with ResNet-18.
As shown in Table~\ref{tab:sparta_trans_data}, we observe that: \ding{182} On both CIFAR-10 and SVHN, ResNet-18 with the transferred \textsc{Sparta} achieves lower top-1 error rates than the one using ReLU under all three attacks. It demonstrates that \textit{pre-trained \textsc{Sparta} on one dataset still works on another dataset, helping CNN get better adversarial robustness than the vanilla ReLU.}
\ding{183} Compared with the standard case where the parameters of ResNet-18 and \textsc{Sparta} are jointly updated during adversarial training, ResNet-18 with transferred \textsc{Sparta} has higher top-1 error rates on adversarial images but lower error rate on clean images. %which is similar with the results in Section~\ref{subsec:trans_CNNs}.

%
%----------------------------------------------------------------------------
\begin{table}[t]
    % % \vspace{-5pt}
	\scriptsize
	\caption{Transferability of \textsc{Sparta} across Datasets. The best results are highlighted.} \label{tab:sparta_trans_data}
	\centering
    \begin{tabular}{l|l|c|c|c|c}
    \toprule
    \multirow{2}{*}{Datasets} & \multirow{2}{*}{ResNet-18} & \multicolumn{3}{c|}{Top-1 error on Adv. Images} & \multirow{2}{*}{\makecell{Top-1 err on\\Clean Img}}\tabularnewline
     &  & PGD-10 & PGD-30 & PGD-50 & \tabularnewline
    \midrule
    \multirow{3}{*}{CIFAR-10} & ReLU & 31.54\% & 68.93\% & 75.64\% & 15.66\%\tabularnewline
    %\cline{2-6} \cline{3-6} \cline{4-6} \cline{5-6} \cline{6-6} 
     & \textsc{Sparta}$_\text{SVHN}$ & \toptwo{30.68\%} & \toptwo{66.93\%}  & \toptwo{74.47\%}  & 15.36\% \tabularnewline
    %\cline{2-6} \cline{3-6} \cline{4-6} \cline{5-6} \cline{6-6} 
     & \textsc{Sparta}$_\text{CIFAR}$ & \topone{29.31\%} & \topone{65.81\%} & \topone{72.55\%} & 15.48\%\tabularnewline
    \midrule
    \multirow{3}{*}{SVHN} & ReLU & 18.27\%  & 63.71\% & 71.42\% & 6.45\% \tabularnewline
    %\cline{2-6} \cline{3-6} \cline{4-6} \cline{5-6} \cline{6-6} 
     & \textsc{Sparta}$_\text{CIFAR}$ & \toptwo{16.59\%} & \toptwo{61.88\%}  & \toptwo{69.45\%} & 6.52\% \tabularnewline
    %\cline{2-6} \cline{3-6} \cline{4-6} \cline{5-6} \cline{6-6} 
     & \textsc{Sparta}$_\text{SVHN}$ & \topone{15.30\%} & \topone{59.29\%}  & \topone{67.29\%} & 7.05\%\tabularnewline
    \bottomrule
    \end{tabular}
    % % \vspace{-15pt}
\end{table}
%----------------------------------------------------------------------------
%

According to the results in Table~\ref{tab:sparta_loc}, we see that: \ding{182} In general, using \textsc{Sparta} at deeper groups helps to achieve better adversarial robustness (\ie, lower top-1 error on adversarial examples) as well as better accuracy (\ie, lower top-1 error rate on clean examples). For example,  ResNet-18-G$_{4}$.B$_2$ achieves the lowest top-1 error on both adversarial and clean examples. \ding{183} Replacing more ReLU layers is not helpful to obtain even better adversarial robustness or accuracy. For example, ResNet-18-G$_{\{1,2,3,4\}}$.B$_{\{1,2\}}$ has eight \textsc{Sparta} layers but obtains higher error rates than G$_{\{1,2,3,4\}}$.B$_2$ with only four \textsc{Sparta} layers. Moreover, replacing the last ReLU layers of all groups, \ie, G$_{\{1,2,3,4\}}$.B$_2$, achieves the best adversarial robustness and the second best accuracy among all variants, indicating the importance of the output activation layers of ResNet groups for adversarial training.

% (2) Adversarial training
% - feature denoising based AT
% - smooth activation AT (swish) \cite{Xie20arXiv}
% - search-based activation for AT (swish) \cite{Ramachandran2017}

% Gradient masking, Obfuscated gradients give a false sense of security. Unfortunately, they are likely to fall under the xxx discuss here. \cite{athalye2018obfuscated}

\subsection{Generalization to SOTA AT Methods}
\label{subsec:generalization}

%
%-----------------------------------------------------------
\begin{table}[t]
    % % \vspace{-5pt}
	\scriptsize
	\caption{Top-1 errors on CIFAR-10 of using TRADES and MART to train ResNet-10 with/without \textsc{Sparta}. Better results are highlighted.} \label{tab:gen_to_ats}
	\centering
    \begin{tabular}{l|c|c|c|c}
    \toprule
    \multirow{2}{*}{ResNet-18} & \multicolumn{2}{c|}{TRADES} & \multicolumn{2}{c}{MART} \tabularnewline
     & ReLU & \textsc{Sparta} & ReLU & \textsc{Sparta} \tabularnewline
    \midrule
    PGD-10    & 41.40\% & \topone{41.19\%} & 41.08\% & \topone{41.05\%} \tabularnewline
    PGD-30    & 69.32\% & \topone{68.76\%} & 66.16\% & \topone{64.98\%}              \tabularnewline
    PGD-50    & 76.02\% & \topone{75.54\%}  & 73.15\% &  \topone{71.68\%}               \tabularnewline
    % Clean Img & 16.30\% & 18.74\% & 17.10\% & 19.50\%\tabularnewline
    \bottomrule
    \end{tabular}
    % % \vspace{-15pt}
\end{table}
%-----------------------------------------------------------
%

In recent years, some works develop more advanced adversarial training (AT) methods \cite{wang2019improving,zhang2019theoretically}. Specifically, Zhang \etal \cite{zhang2019theoretically} design a new formulation of adversarial training based on the trade-off between robustness and accuracy (\ie, TRADES).  
Wang \etal \cite{wang2019improving} propose a novel adversarial training algorithm (\ie, misclassification-aware adversarial training (MART)) according to the influence of misclassified and correctly classified examples. 
Note that, the two methods are improved adversarial training approaches while our work is to enhance the basic network component (\ie, ReLU activation) for the standard adversarial training. 
An interesting problem is whether our method still benefits the advanced AT methods (\ie, TRADES and MART) or not.
To validate this problem, we can use TRADES and MART to train the networks equipped with \textsc{Sparta} to see whether our method can generalize to these new adversarial training approaches and further improve the robustness.

To this end, we use the official implementation of TRADES and MART to train the ResNet-10 with/without \textsc{Sparta} on CIFAR-10 and compare their results under clean images and PGD-10 attack. Note that, TRADES and MART are implemented in PyTorch while we implement \textsc{Sparta} in TensorFlow. For a fair comparison, we follow the setups of TRADES and MART and re-implement the ResNet-18 with \textsc{Sparta} in PyTorch. We report the results in \tableref{tab:gen_to_ats} and see that: ResNet-18s equipped with \textsc{Sparta} could achieve lower top-1 errors than the models with ReLU when we conduct training via TRADES and MART, which demonstrate that our method can be generalized to more advanced adversarial training techniques.
\\

%----------------------------------------------------------
%----------------------------------------------------------
\section{Conclusions}\label{sec:concl}
We have proposed a novel activation function, named \textsc{Sparta}, which is designed to be spatially attentional and adversarially robust. It enables CNNs to achieve higher robustness, without hurting accuracy on clean examples, than the ones based on the state-of-the-art activation functions. We have investigated the relationships between \textsc{Sparta} and the state-of-the-art search-based activation, \ie, Swish, and feature denoising method, providing insights about the advantages of our method. Furthermore, comprehensive evaluation presents two important properties of our method: \emph{superior transferability across CNNs} and \emph{superior transferability across datasets}, confirming \textsc{Sparta}'s important properties of flexibility and versatility.

\backmatter

\bibliography{sn-bibliography}% common bib file

\end{document}